\documentclass[11pt]{article} 
\usepackage[left=1.2in, top=1in, bottom=1in, right=1.2in]{geometry}
\usepackage{times}
\usepackage{subcaption}
\usepackage{caption}

\usepackage[utf8]{inputenc} 
\usepackage[T1]{fontenc}    
\usepackage{hyperref}       
\usepackage{url}            
\usepackage{booktabs}       
\usepackage{amsfonts,amsthm,amsmath,amssymb}       
\usepackage{nicefrac}       
\usepackage{microtype}      
\usepackage{xcolor}         
\usepackage{txfonts}
\usepackage{graphicx}
\usepackage{wrapfig,bm,comment,color}
\usepackage{breakurl,epsfig,epsf,fmtcount,semtrans,multirow,boldline}
\usepackage{tcolorbox}
\tcbuselibrary{skins}
\usepackage{tikz}
\usepackage{enumerate}
\usepackage{enumitem}

\definecolor{darkred}{RGB}{150,0,0}
\definecolor{darkgreen}{RGB}{0,150,0}
\definecolor{darkblue}{RGB}{0,0,200}
\hypersetup{colorlinks=true, linkcolor=darkred, citecolor=darkgreen, urlcolor=darkblue}

\newtheorem{theorem}{Theorem}

\newtheorem{claim}{Claim}
\newtheorem{assumption}{Assumption}

\newtheorem{lemma}{Lemma}
\newtheorem{corollary}{Corollary}

\newtheorem{definition}{Definition}

\newcommand{\beq}{\begin{equation}}
\newcommand{\ba}{\begin{align}}
\newcommand{\ea}{\end{align}}

\newcommand{\eeq}{\end{equation}}
  
\newcommand{\vct}[1]{\bm{#1}}
\newcommand{\tr}[1]{\texttt{tr}\left(#1\right)}
\newcommand{\mtx}[1]{\bm{#1}}

\newcommand{\hb}{\vct{h}}

\newcommand{\st}{\star}
\newcommand{\h}{\boldsymbol{h}}

\newcommand{\la}{\lambda}
\newcommand{\bla}{\boldsymbol{\lambda}}
\newcommand{\A}{{\mtx{A}}}

\newcommand{\B}{{\mtx{B}}}

\newcommand{\Ub}{{\mtx{U}}}

\newcommand{\Sb}{{{\mtx{S}}}}

\newcommand{\Gb}{{\mtx{G}}}

\newcommand{\Lc}{{\cal{L}}}

\newcommand{\Nc}{{\cal{N}}}

\newcommand{\Dc}{{\cal{D}}}

\newcommand{\Pb}{{\mtx{P}}}
\newcommand{\Pt}{{\tilde{\mtx{P}}}}
\newcommand{\Pk}{\mtx{P}_k}
\newcommand{\Pq}{\mtx{P}_q}
\newcommand{\Pkl}{\mtx{P}_{k,\ell}}
\newcommand{\Pl}{\mtx{P}_{\ell}}
\newcommand{\Pql}{\mtx{P}_{q,\ell}}

\newcommand{\Eb}{{\mtx{E}}}

\newcommand{\Gc}{{\cal{G}}}

\newcommand{\bmu}{{\boldsymbol{\mu}}}
\newcommand{\Db}{{\mtx{D}}}

\newcommand{\db}{{\vct{d}}}
\newcommand{\dbb}{{\vct{\bar{d}}}}
\newcommand{\onebb}{{\mtx{1}}}
\newcommand{\zerobb}{{\mtx{0}}}
\newcommand{\Iden}{{\mtx{I}}}
\newcommand{\M}{{\mtx{M}}}
\newcommand{\z}{{\vct{z}}}

\newcommand{\bom}{\boldsymbol{\omega}}
\newcommand{\bOm}{\boldsymbol{\Omega}}

\newcommand{\bal}{{\boldsymbol{\alpha}}}
\newcommand{\bt}{{\boldsymbol{\beta}}}

\newcommand{\Nn}{\mathcal{N}}
\newcommand{\vb}{\vct{v}}

\newcommand{\diag}[1]{{\text{diag}}(#1)}
\newcommand{\cb}{\mtx{c}}
\newcommand{\cbb}{\bar{\mtx{c}}}
\newcommand{\w}{\vct{w}}
\newcommand{\ob}{\mtx{o}}
\newcommand{\Ob}{\mtx{O}}
\newcommand{\li}{\left<}
\newcommand{\ri}{\right>}
\newcommand{\s}{\vct{s}}
\newcommand{\ab}{\boldsymbol{a}}

\newcommand{\ub}{{\vct{u}}}

\newcommand{\g}{{\vct{g}}}

\newcommand{\Z}{\mtx{Z}}

\newcommand{\kb}{\vct{k}}
\newcommand{\bSi}{\boldsymbol{\Sigma}}

\newcommand{\x}{\vct{x}}
\newcommand{\rb}{\vct{r}}
\newcommand{\y}{\vct{y}}
\newcommand{\ym}[2]{{y_{#1}^{#2}}}

\newcommand{\gm}[2]{{\g_{#1}^{#2}}}

\newcommand{\zm}[2]{{\z_{#1}^{#2}}}
\newcommand{\obm}[2]{{\ob_{#1}^{#2}}}

\newcommand{\yhm}[2]{{\hat y_{#1}^{#2}}}
\newcommand{\ybm}[2]{{\bar y_{#1}^{#2}}}
\newcommand{\Yhm}[2]{{\hat \y_{#1}^{#2}}}

\newcommand{\btm}[2]{{\bt_{#1,#2}}}
\newcommand{\nablam}[2]{{\nabla_{#1}^{#2}}}

\newcommand{\W}{\mtx{W}}
\newcommand{\Wk}{{\mtx{W}_k}}
\newcommand{\Wq}{{\mtx{W}_q}}
\newcommand{\Wv}{{\mtx{W}_v}}

\newcommand{\Wc}{{\cal{W}}}

\newcommand{\X}{{\mtx{X}}}
\newcommand{\Xt}{{\tilde{\mtx{X}}}}

\newcommand{\Rb}{{\mtx{R}}}

\newcommand{\qb}{{\vct{q}}}

\newcommand{\R}{\mathbb{R}}

\newcommand{\E}{{\mathbb{E}}}

\newcommand{\nn}{\nonumber}

\newcommand{\tn}[1]{\left\|{#1}\right\|_{\ell_2}}

\newcommand{\order}[1]{{\cal{O}}(#1)}

\newcommand{\Wt}{\tilde{\mtx{W}}}

\newcommand{\Xb}{\mtx{\bar{X}}}

\newcommand{\Mb}{\bar{\M}}

\newcommand{\yb}{\bar{\y}}

\newcommand{\att}{{\texttt{ATT}}}
\newcommand{\gla}{{\texttt{GLA}}}
\newcommand{\wpgd}{{\texttt{WPGD}}}

\usepackage[textsize=tiny,textwidth=3cm]{todonotes}

\usepackage[normalem]{ulem}

\newcommand{\SO}[1]{\textcolor{blue}{[SO: #1]}}

\DeclareMathOperator*{\argmin}{argmin}

\usepackage{hyperref}
\usepackage{url}
\usepackage{natbib}

\title{Gating is Weighting: Understanding Gated Linear Attention through In-context Learning}

\author{Yingcong Li$^\star$$^1$ \quad Davoud Ataee Tarzanagh$^\star$$^2$ \\ Ankit Singh Rawat$^3$ \quad Maryam Fazel$^4$ \quad Samet Oymak$^1$}
\date{}
\begin{document}
\addtocontents{toc}{\protect\setcounter{tocdepth}{0}}

\maketitle
{\let\thefootnote\relax\footnotetext{$^1$ University of Michigan, \texttt{\{yingcong,oymak\}@umich.edu}. $^2$ University of Pennsylvania, \texttt{tarzanaq@upenn.edu}. \\$^3$ Google Research NYC, \texttt{ankitsrawat@google.com}. $^4$ University of Washington, \texttt{mfazel@uw.edu}. $^\star$ Equal contribution.}}

\begin{abstract} 
Linear attention methods offer a compelling alternative to softmax attention due to their efficiency in recurrent decoding. Recent research has focused on enhancing standard linear attention by incorporating gating while retaining its computational benefits. Such Gated Linear Attention (GLA) architectures include competitive models such as Mamba and RWKV. In this work, we investigate the in-context learning capabilities of the GLA model and make the following contributions. We show that a multilayer GLA can implement a general class of Weighted Preconditioned Gradient Descent (WPGD) algorithms with data-dependent weights. These weights are induced by the gating mechanism and the input, enabling the model to control the contribution of individual tokens to prediction. To further understand the mechanics of this weighting, we introduce a novel data model with multitask prompts and characterize the optimization landscape of learning a WPGD algorithm. Under mild conditions, we establish the existence and uniqueness (up to scaling) of a global minimum, corresponding to a unique WPGD solution. Finally, we translate these findings to explore the optimization landscape of GLA and shed light on how gating facilitates context-aware learning and when it is provably better than vanilla linear attention.
\end{abstract}

\section{Introduction}

The Transformer \citep{vaswani2017attention} has become the de facto standard for language modeling tasks. The key component of the Transformer is the self-attention mechanism, which computes softmax-based similarities between all token pairs. Despite its success, the self-attention mechanism has quadratic complexity with respect to sequence length, making it computationally expensive for long sequences. To address this issue, a growing body of work has proposed near-linear time approaches to sequence modeling. The initial approaches included linear attention and state-space models, both achieving $O(1)$ inference complexity per generated token, thanks to their recurrent form. While these initial architectures typically do not match softmax attention in performance, recent recurrent models such as Mamba~\citep{gu2023mamba}, mLSTM~\citep{beck2024xlstm}, GLA Transformer~\citep{yang2023gated}, and RWKV-6~\citep{peng2024eagle} achieve highly competitive results with the softmax Transformer. Notably, as highlighted in \citet{yang2023gated}, these architectures can be viewed as variants of \textit{gated linear attention} (GLA), which incorporates a gating mechanism within the recurrence of linear attention.

Given a sequence of tokens $(\z_i)_{i=1}^{n+1} \in \R^{d+1}$ and the associated query, key, and value embeddings $(\qb_i,\kb_i,\vb_i)_{i=1}^{n+1} \subset \mathbb{R}^{d+1}$, with $d+1$ being the embedding dimension, the GLA recurrence is given by
\begin{equation}\tag{GLA}
\label{eq:GLA}
\begin{split}    
&\Sb_{i} = \Gb_i \odot \Sb_{i-1} + \vb_i \kb_i^\top,\quad\text{and}\\
&\ob_i = \Sb_{i} \qb_i, \quad   i  \in \left\{1,\ldots, n+1\right\}.
\end{split}
\end{equation}
Here, $\Sb_i \in \mathbb{R}^{(d+1) \times (d+1)}$ represents the \textit{2D state variable} with $\Sb_0 = \mathbf{0}$; $\ob_i \in \mathbb{R}^{d+1}$ represents the $i$th output token \footnote{The sequence length and embedding dimension are set to \( n+1 \) and \( d+1 \) per the prompt definition in \eqref{input}.}; and the \textit{gating variable} 
$\Gb_i := G(\z_i) \in \mathbb{R}^{(d+1) \times (d+1)}$ is applied to the state $\Sb_{i-1}$ through the Hadamard product $\odot$ and $G$ represents the gating function. When $\Gb_i$ is a matrix of all ones, \eqref{eq:GLA} reduces to causal linear attention~\citep{katharopoulos2020transformers}.

The central objective of this work is to enhance the mathematical understanding of the GLA mechanism. In-context learning (ICL), one of the most remarkable features of modern sequence models, provides a powerful framework to achieve this aim. ICL refers to the ability of a sequence model to implicitly infer functional relationships from the demonstrations provided in its context window \citep{brown2020language,min2022rethinking}. It is inherently related to the model's ability to emulate learning algorithms. Notably, ICL has been a major topic of empirical and theoretical interest in recent years.  More specifically, a series of works have examined the approximation and optimization characteristics of linear attention, and have provably connected linear attention to the preconditioned gradient descent algorithm \citep{von2023transformers,ahn2024transformers,zhang2024trained}. Given that the GLA recurrence in \eqref{eq:GLA} has a richer design space, this leads us to ask: 
\begin{quote}
\begin{center}    
\textit{{What learning algorithm does GLA emulate in ICL?\\How does its optimization landscape behave in multi-task prompting?}}
\end{center}
\end{quote}
\noindent\textbf{Contributions:} The GLA recurrence in \eqref{eq:GLA} enables the sequence model to weight past information in a data-dependent manner through the gating mechanism $(\Gb_i)_{i=1}^{n}$. Building on this observation, we demonstrate that a GLA model can implement a \emph{data-dependent Weighted Preconditioned Gradient Descent (WPGD)} algorithm. Specifically, a one-step version of this algorithm with scalar gating, where all entries of $\Gb_i$ are identical, is described by the prediction:
\begin{align}
\hat{y} = \x^\top \Pb \X^\top(\y \odot \bom).\label{wpgd}
\end{align}
Here, $\X \in \R^{n \times d}$ is the input feature matrix; $\y \in \R^n$ is the associated label vector; $\x \in \R^d$ represents the test/query input to predict; $\Pb \in \R^{d \times d}$ is the preconditioning matrix; and $\bom \in \R^n$ weights the individual samples. When $\bom$ is fixed, we drop ``data-dependent'' and simply refer to this algorithm as the WPGD algorithm. However, for GLA, the weighting vector $\bom$
depends on the data through recursive multiplication of the gating variables. Building on this formalism, we make the following specific contributions:
\begin{enumerate}[label=$\diamond$]
\item \textbf{ICL capabilities of GLA (\S\ref{sec plain wpgd}):} Through constructive arguments, we demonstrate that a multilayer GLA model can implement data-dependent WPGD iterations, with weights induced by the gating function. This construction sheds light on the role of causal masking and the expressivity distinctions between scalar- and vector-valued gating functions.
\item \textbf{Landscape of {{one}}-step WPGD (\S\ref{sec:optim gd}):} The GLA$\Leftrightarrow$WPGD connection motivates us to ask: \emph{How does WPGD weigh demonstrations in terms of their relevance to the query?} To address this, we study the fundamental problem of learning an optimal WPGD algorithm: Given a tuple $(\X,\y,\x,y)\sim \Dc$, with $y\in\R$ being the label associated with the query, we investigate the population risk minimization:
\begin{equation}\label{eqn:wpgd:risk}
\begin{split}
&\Lc_\wpgd^\st := \min_{\Pb {\in \mathbb{R}^{d \times d}},\bom \in \R^n}~\Lc_\wpgd(\Pb,\bom), ~\\
&\text{where}~~\Lc_\wpgd(\Pb,\bom)=\E_{\Dc}\left[\left(y-\x^\top \Pb \X^\top(\y\odot\bom)\right)^2\right].
\end{split}
\end{equation}
As our primary mathematical contribution, we characterize the loss landscape under a general multitask data setting, where the tasks associated with the demonstrations $(\X,\y)$ have varying degrees of correlation to the target task $(\x,y)$. We carefully analyze this loss landscape and show that, under mild conditions, there is a unique global minimum $(\Pb,\bom)$ up to scaling invariance, and the associated WPGD algorithm is also unique.

\item \textbf{Loss landscape of {{one}}-layer GLA (\S\ref{gla sec}):} The landscape is highly intricate due to the recursively multiplied gating variables. We show that learning the optimal GLA layer can be connected to solving \eqref{eqn:wpgd:risk} with a constraint $\bom\in\Wc$, where the restriction $\Wc$ is induced by the choice of gating function and input space. Solidifying this connection, we introduce a multitask prompt model under which we characterize the loss landscape of GLA and the influence of task correlations. Our analysis and experiments reveal insightful distinctions between linear attention, GLA with scalar gating, and GLA with vector-valued gating.
\end{enumerate}

\section{Problem setup}\label{sec:setup}
\emph{Notations.}
$\mathbb{R}^d$ is the $d$-dimensional real space, with $\mathbb{R}^d_{+}$ and $\mathbb{R}^d_{++}$ as its positive and strictly positive orthants. The set $[n]$ denotes $\{1, \dots, n\}$. Bold letters, e.g., $\ab$ and $\A$, represent vectors and matrices. The identity matrix of size $n$ is denoted by $\Iden_n$. The symbols $\onebb$ and $\zerobb$ denote the all-one and all-zero vectors or matrices of proper size, such as $\onebb_d \in \mathbb{R}^d$ and $\onebb_{d\times d} \in \mathbb{R}^{d\times d}$. The subscript is omitted when the dimension is clear from the context. The Gaussian distribution with mean $\bmu$ and covariance $\bSi$ is written as $\mathcal{N}(\bmu,\bSi)$. The Hadamard product (element-wise multiplication) is denoted by $\odot$, and Hadamard division (element-wise division) is denoted by $\oslash$. Given any $\ab_{i+1}, \dots, \ab_j \in \mathbb{R}^d$, we define $\ab_{i:j} = \ab_{i+1} \odot \cdots \odot \ab_j$ for $i < j$, and $\ab_{i:i} = \onebb_d$.

The objective of this work is to develop a theoretical understanding of GLA through ICL. The optimization landscape of standard linear attention has been a topic of significant interest in the {ICL} literature~\citep{ahn2024transformers,li2024fine}. Following these works, we consider the input prompt
\begin{align}
\Z=[\z_1~~~\dots~~~\z_n~~~\z_{n+1}]^\top=
\begin{bmatrix}
\x_1 & \dots & \x_n & \x_{n+1} \\
y_1 & \dots & y_n & 0
\end{bmatrix}^\top
\in\R^{(n+1)\times(d+1)},\label{input}
\end{align}
where tokens encode the input-label pairs $(\x_i,y_i)_{i=1}^{n}\subset\R^d\times\R$. 

We aim to enable ICL by training a sequence model $F: \R^{(n+1) \times (d+1)} \to \R
$ that predicts the label $y:=y_{n+1}$ associated with the query $\x:=\x_{n+1}$. 
This model will utilize the demonstrations $(\x_i,y_i)_{i=1}^n$ to infer the mapping between $\x$ and $y$. Assuming that the data is distributed as $(y,\Z)\sim \Dc$, the ICL objective is defined as
\begin{align}
\Lc(F)=\E_{\Dc}\left[\left(y-F(\Z)\right)^2\right].\label{ICL loss}
\end{align}

\paragraph*{Linear attention and shared-task distribution.}Central to our paper is the choice of the function class $F$. When $F$ is a linear attention model, 
the prediction denoted by $\hat{y}_{F}$ takes the form  
\begin{align*}
\hat{y}_{F} = F(\Z)=\z_{n+1}^\top \W_q \W_k^\top   \Z^\top \Z \W_v \h,    
\end{align*}
where $\W_{k}, \W_{q}, \W_{v} \in \mathbb{R}^{(d+1) \times (d+1)}$ are attention parameters, and $\h \in \mathbb{R}^{d+1}$ is the linear prediction head. 

To motivate our subsequent discussion on multitask distributions, we begin by reviewing a simplified \emph{shared-task distribution}. Specifically, we let $\bt \sim \mathcal{N}(\zerobb, \bSi_\bt)$, $\x_i$ be i.i.d. with $\x_i \sim \mathcal{N}(\zerobb, \bSi_{\x})$, and $y_i \sim \mathcal{N}(\bt^\top \x_i, \sigma^2)$, where $\sigma \geq 0$ denotes the noise level. Although this is not our primary modeling assumption, under this shared-task distribution, it has been shown \citep{von2023transformers,ahn2024transformers,zhang2024trained} that the optimal one-layer linear attention prediction coincides with one-step preconditioned gradient descent (PGD). In particular, given the data distribution above, we have:
\begin{align}\label{PGD pred}
\hat y_F=\x^\top\hat\bt, \quad\text{where}\quad \hat\bt = \Pb^\star\X^\top \y,
\end{align}
and
\begin{align}
\Pb^\star:=\argmin_{\Pb \in \mathbb{R}^{d \times d}}~\E\left[\left(y-\x^\top \Pb\X^\top \y\right)^2\right]\quad\text{with}\quad \X:= \begin{bmatrix} \x_1~\cdots~\x_n\end{bmatrix}^\top ~~\text{and}~~ \y:=\begin{bmatrix}y_1~\cdots~y_n\end{bmatrix}^\top.\label{PGD loss}
\end{align}

\paragraph*{Gated linear attention and multitask prompting.} 
Given the input prompt $\Z$ as in \eqref{input}, let $(\qb_i, \kb_i,\vb_i)= (\W_q^\top\z_i, \W_k^\top\z_i,\W_v^\top\z_i)$ be the corresponding query, key,
and value embeddings.   The output of \textit{causal} linear attention at time $i$ can be computed using \eqref{eq:GLA} with $\Gb_i = \onebb$ for all  $i  \in [n+1]$.  This recurrent form implies that linear attention has $O(d^2)$ cost, that is independent of sequence length $n$, to generate per-token. 
GLA follows the same structure as linear attention but with a gating mechanism {(i.e., $\Gb_i \neq \onebb$ for some  $i  \in [n+1]$} ), which equips the model with the option to pass or supress the history.  As discussed in \cite{yang2023gated}, the different choices of the gating correspond to different popular recurrent architectures such as Mamba \citep{gu2023mamba}, Mamba2 \citep{dao2024transformers}, RWKV \citep{peng2024eagle}, etc.

We will show that GLA can weigh the context window through gating, thus, its capabilities are linked to the WPGD algorithm described in \eqref{eq:wpgd}. This will in turn facilitate GLA to effectively learn \emph{multitask prompt distributions} described by $y_i\sim\Nn(\bt_i^\top\x_i,\sigma^2)$ with $\bt_i$'s not necessarily identical. 
\section{What gradient methods can GLA emulate?}\label{sec plain wpgd}
In this section, we investigate the ICL capabilities of {{GLA}} and show that under suitable 
{instantiations of model weights}, GLA can implement \textit{data-dependent} WPGD. 

\subsection{{GLA as a data-dependent WPGD predictor}}  

\paragraph*{Data-Dependent WPGD.}  
Given $\X$ and $\y$ as defined in \eqref{PGD loss}, consider the weighted least squares objective $\Lc(\bt)=\sum_{i=1}^n \omega_i\cdot (y_i-\bt^\top \x_i)^2$ with weights $\bom=[\omega_1,\omega_2,~\cdots,~\omega_n]^\top\in\R^n$. To optimize this, we use gradient descent (GD) starting from zero initialization, $\bt_0 = \zerobb$ with a step size of $\eta = 1/2$. One step of standard GD is given by
\begin{align*}
    \bt_1 = \bt_0 - \eta\nabla\Lc(\bt_0)=\sum_{i=1}^n \omega_i \cdot \x_i y_i=\X^\top (\bom\odot\y).
\end{align*}  
Given a test/query feature $\x$, the corresponding prediction is $\hat{y} = \x^\top \hat{\bt}$ where $ \hat{\bt} = \bt_1$. Additionally, if we were using PGD with a {preconditioning matrix $\Pb\in\R^{d\times d}$, $\bt_0 = \zerobb$, and $\eta = 1/2$, then a single iteration results in}
\begin{equation}\label{eqn:gd:wpgd-basic}
\hat{y} = \x^\top \hat{\bt}, \quad \text{where} \quad {\hat{\bt}  =\bt_0 -  \eta \Pb \nabla\Lc(\bt_0)}= \Pb \X^\top (\bom\odot\y).
\end{equation}
Above is the basic {\emph{sample-wise} WPGD} predictor which weights individual datapoints, in contrast to the PGD predictor as presented in \eqref{PGD pred}. It turns out, \emph{vector-valued gating} can facilitate a more general estimator which weights individual coordinates. To this aim,  we introduce an extension 
as follows: Let $\Pb_1, \Pb_2 \in \R^{d \times d}$ denote the preconditioning matrices, and let $\bOm \in \R^{n \times d}$ denote the \emph{vector-valued weighting} matrix. Note that $\bOm$ is a weight matrix that enables coordinate-wise weighting. Throughout the paper, we use $\bom$ and $\bOm$ to denote sample-wise and coordinate-wise weighting parameters, corresponding to scalar and vector gating strategies, respectively. Then given $\bOm$, we can similarly define
\begin{subequations}\label{eq:wpgd}
\begin{equation}
\bt_{\text{1}}^{\text{gd}}(\Pb_1, \Pb_2, \bOm) := \Pb_2 (\X \Pb_1 \odot \bOm)^\top \y \label{eq:wpgd:pred}
\end{equation} 
as one-step of (generalized) WPGD. Its corresponding prediction on a test query $\x$ is:
\begin{equation} 
\hat{y} = \x^\top \hat{\bt}, \quad \text{where} \quad \hat{\bt} = \bt_{\text{1}}^{\text{gd}}(\Pb_1, \Pb_2, \bOm).
\end{equation} 
\end{subequations}
We note that by removing the weighting matrix $\bOm$, \eqref{eq:wpgd} reduces to standard PGD (cf. \eqref{PGD pred}) and setting $\bOm=\bom\onebb_d^\top$ reduces to sample-wise WPGD (cf. \eqref{eqn:gd:wpgd-basic}).
\cite{li2024fine} has demonstrated that H3-like models implement one-step sample-wise WPGD, and they focus on the shared-task distribution where $\bt_i \equiv \bt$ for all $i \in [n]$. In contrast, our work considers a more general data setting where tasks within an in-context prompt are not necessarily identical.  

We introduce the following model constructions under which we establish the equivalence between GLA (cf. \eqref{eq:GLA}) and WPGD (cf. \eqref{eq:wpgd}), where the weighting matrix is induced by the input data and the gating function. Inspired by prior works \citep{von2023transformers,ahn2024transformers}, we consider the following restricted attention matrices:
\begin{align}
\Wk &= \begin{bmatrix}
    \Pk & \zerobb \\
    \zerobb & 0
\end{bmatrix}, \quad
\Wq = \begin{bmatrix}
    \Pq & \zerobb \\
    \zerobb & 0
\end{bmatrix}, \quad \text{and} \quad
\Wv = \begin{bmatrix}
    \zerobb_{d\times d} & \zerobb \\
    \zerobb & 1
\end{bmatrix}, \label{att mtx res}
\end{align}
where $\Pk, \Pq \in \R^{d \times d}$. Note that the $(d+1, d+1)$-th entry of $\Wv$ is set to one for simplicity. More generally, this entry can take any nonzero value, e.g., $v \in \R$. Parameterizing $\Wq$ with $\Pq / v$ produces the same output as in \eqref{att mtx res}.

\begin{theorem}\label{thm:gla=wpgd} Recall \eqref{eq:GLA} with input sequence $\Z =[\z_1~~\dots~~\z_n~~\z_{n+1}]^\top$ defined in \eqref{input},  the gating  $G(\z_i)=\Gb_i\in\R^{(d+1)\times (d+1)}$, and outputs  $(\ob_{i})_{i=1}^{n+1}$. 
Consider model construction in \eqref{att mtx res} and take the last coordinate of the last token output  denoted by $(\ob_{n+1})_{d+1}$ as a prediction. Then, we have 
\[
f_{\gla}(\Z):=(\ob_{n+1})_{d+1}
=\x^\top\hat\bt,\quad\textnormal{where}\quad\hat\bt=\bt_{\text{1}}^{\text{gd}}(\Pk,\Pq,\bOm).
\]
Here, $\bt_{\text{1}}^{\text{gd}}(\cdot)$ is a one-step WPGD feature predictor defined in \eqref{eq:wpgd:pred}; $\Pb_k$ and $\Pb_q$ correspond to attention weights following \eqref{att mtx res}; and
$\bOm=[\g_{1:n+1}~\g_{2:n+1}~\cdots~\g_{n:n+1}]^\top\in\R^{n\times d}$, where $\g_{i:n+1},i\in[n]$ is given by
\begin{align}
\g_{i:n+1}:= {\g_{i+1}\odot \g_{i+2}\cdots \g_{n+1}} \in\R^d,\quad\textnormal{and}\quad\Gb_i=\begin{bmatrix}
    *&*\\
    \g_i^\top&*
\end{bmatrix}
.
\label{weighting}
\end{align}
\end{theorem}
Here, and throughout, we use $*$ to fill the entries of the matrices that do not affect the final output, and based on the {model construction} given in \eqref{att mtx res}, these entries can be assigned any value.

Observe that, crucially, since $\g_i$ is associated with $\z_i$, $\z_i$ influences the weighting of $\z_{j}$ {for all $j <i$}. We defer the proof of Theorem~\ref{thm:gla=wpgd} to the Appendix~\ref{app:1 layer}. It is noticeable that only $d$ of the total $(d+1)^2$ entries in each gating matrix $\Gb_i$ are useful due to the model construction presented in \eqref{att mtx res}.   However, if we relax the weight restriction, e.g., $\Wv=[\zerobb_{(d+1)\times d}~~\ub]^\top$, where $\ub\in\R^{d+1}$, then the weighting matrix $\bOm$ in Theorem~\ref{thm:gla=wpgd} is associated with all rows of the $\Gb_i$ matrices. We defer to Eqn. \eqref{eqn:allrow} and  the discussion in Appendix~\ref{app:1 layer}. 

\subsection{Capabilities of multi-layer GLA}
\citet{ahn2024transformers} demonstrated that, with appropriate construction, an $L$-layer linear {attention model}
performs $L$-step preconditioned GD on the dataset $(\x_i, y_i)_{i=1}^n$ provided within the prompt. 
In this work, we study multi-layer GLA and analyze the associated algorithm class it can emulate. 
{It is worth mentioning that \citet{ahn2024transformers} does not consider \textit{causal masking} which is integral to multilayer GLA due to its recurrent nature described in \eqref{eq:GLA}. Our analysis will capture the impact of gating and causal mask through $n$ separate GD trajectories that are coupled.}

\paragraph*{Multi-layer GLA.} For any input prompt $\Z\in\R^{(n+1)\times(d+1)}$, let $\gla(\Z):=[\ob_1~~\ob_2~~\cdots~~\ob_{n+1}]^{\top}\in\R^{(n+1)\times(d+1)}$ denote its GLA output, as defined in \eqref{eq:GLA}.  
We consider an $L$-layer GLA model as follows:  
For each layer $\ell\in[L]$, let $\Z_\ell$ denote its input, where we set $\Z_1=\Z$, and let $\W_{k,\ell},\W_{q,\ell},\W_{v,\ell}$ denote the key, query, and value matrices for layer $\ell$, respectively. After applying a residual connection, the input of the $\ell$-th layer is given by
\begin{align}\label{gla multi}
    \Z_\ell:=\Z_{\ell-1}+\gla_{\ell-1}(\Z_{\ell-1}). 
\end{align}
Here, $\gla_\ell(\cdot)$ denotes the output of the $\ell$-th GLA layer, associated with the attention matrices $(\W_{q,\ell},\W_{k,\ell},\W_{v,\ell})$ for all $\ell\in[L]$. In the following, we focus on the $(d+1)$-th entry of each token's output at each layer (after the residual connection), denoted by $o_{i,\ell} := [\Z_\ell+\gla_\ell(\Z_\ell)]_{i,d+1}$ for all $i \in [n+1]$ and $\ell\in[L]$.

\begin{theorem}\label{lemma:multi layer}
Consider an $L$-layer GLA  defined in \eqref{gla multi}, 
where  $\W_{k,\ell}$, $\W_{q,\ell}$ in the $\ell$'th layer parameterized by $\Pkl, -\Pql \in \R^{d \times d}$, for all  $\ell \in [L]$ following \eqref{att mtx res}. Let the gating be a function of the features, e.g., $\Gb_i =G(\x_i)$, and define $\bOm$ as in Theorem~\ref{thm:gla=wpgd}. 
Additionally, denote the masking as $\M_i = \begin{bmatrix}
    \Iden_i & \zerobb\\
    \zerobb & \zerobb
\end{bmatrix} \in \R^{n \times n}$, and  let $\hat\bt_0=\btm{i}{0} = \zerobb$ for all $i \in [n]$. Recall that $ o_{i,\ell}$ for all  $i \in [n+1]$ and $\ell\in[L]$ stands for the last entry of the $i$'th token output at the $\ell$'th layer. We have 
\begin{itemize}
    \item For all $i \leq n$, $ o_{i,\ell} = y_i - \x_i^\top \bt_{i,\ell}$, where $\bt_{i,\ell} = \bt_{i,\ell-1} {-\Pql} \left(\nabla_{i,\ell} \oslash \g_{i:n+1}\right)$;
    \item $ o_{n+1,\ell} = -\x^\top \hat\bt_{\ell}$, where $\hat\bt_{\ell} = (1 - \alpha_{\ell}) \hat\bt_{\ell-1} -\Pql \left(\nabla_{n,\ell} \oslash \g_{n+1}\right)$, and $\alpha_\ell = \x^\top \Pql \Pkl^\top \x$.
\end{itemize}
Here, letting $\B_{\ell} = [\btm{1}{\ell}~\cdots~\btm{n}{\ell}]^\top$, $\X_{\ell} = \X\Pb_{k,\ell} \odot \bOm$, 
and $\y_\ell = \textnormal{diag}(\X\B_\ell^\top)$, we define 
\[
\nabla_{i,\ell} = \X_{\ell}^\top \M_i \left(\y_\ell-\y\right).
\]
\end{theorem}

We defer the proof of Theorem~\ref{lemma:multi layer} to the Appendix~\ref{app:multi layer}. Theorem~\ref{lemma:multi layer} states that $L$-layer GLA \eqref{gla multi}  implements $L$ steps of WPGD with gradient in a recurrent form and additional weight decay $\alpha_\ell$. To recap, given data $(\X, \y)$ and prediction $\hat\bt$, the gradient with respect to the squared loss takes the form $\X^\top(\X\hat\bt - \y)$, up to some constant $c$. In comparison, $\Pql\left(\nabla_{i,\ell}\oslash\g_{i:n+1}\right)$ similarly acts as a gradient but incorporates layer-wise feature preconditioners ($\Pql, \Pkl$), data weighting ($\bOm$), and causality ($\g_{i:n+1}, \M_i$). Here, $\M_i$ represents causal masking, ensuring that at time $i$, only inputs from $j \leq i$ are used for prediction. Notably, the recurrent structure of GLA allows the gating mechanism to apply context-dependent weighting strategies.

To simplify the theorem statement, we assume that the gating function depends only on the input feature, e.g., $\Gb_i = G(\x_i)$, ensuring that the corresponding data-dependent weighting is uniform across all layers. This assumption is included solely for clarity in the theorem statement, and if the gating function varies across layers or depends on the entire $(d+1)$-dimensional token rather than just the feature $\x_i$, each layer has its own gating $\bOm_\ell$ and $\X_\ell=\X\Pkl\odot\bOm_{\ell}$. Note that our inclusion of the additional term $\alpha_\ell$ captures the influence of the last token's output on the next layer’s prediction. Based on the above $L$-layer GLA result, we have the following corollary for multi-layer linear attention network with causal mask in each layer.


\begin{corollary}\label{corol multi}
    Consider an $L$-layer linear attention model with causal mask and residual connection in each layer. Let $\ell$'th layer follows the weight construction in \eqref{att mtx res} with $\W_{k,\ell},\W_{q,\ell}$ parameterized by $\Pkl,{-\Pql}$ as in Theorem~\ref{lemma:multi layer} and define $\Pb_\ell:={\Pql\Pkl^\top}$, for $\ell\in[L]$. Let $\hat\bt_0=\zerobb$. Then, the $(d+1)$'th entry of the last token output of each layer satisfies:
    \begin{align*}
         o_{n+1,\ell}=-\x^\top\hat\bt_{\ell}, \quad \text{where}\quad \hat\bt_{\ell}=(1-\alpha_{\ell})\hat\bt_{\ell-1}{-\Pl}\X^\top(\y_{\ell}-\y).
    \end{align*}
Here, we define $\alpha_\ell=\x^\top\Pl\x$ and $\y_\ell$ follows the same definition as in Theorem~\ref{lemma:multi layer}.
\end{corollary}

Note that Corollary~\ref{corol multi} generalizes \cite[Lemma 1]{ahn2024transformers} by extending it to causal attention. When considering full attention (i.e., without applying the causal mask), we have $\y_\ell = \X\hat\bt_\ell$, and the expression $\X^\top(\y_\ell-\y) = \X^\top(\X\hat\bt_\ell - \y)$ corresponds to the standard gradient of the squared loss. Furthermore, if we disregard the impact of the last token (e.g., by incorporating the matrix $M$ as in Eq. (3) of \cite{ahn2024transformers}), then $\alpha_\ell = 0$. These results are also consistent with \cite{ding2023causallm}, which demonstrate that causal masking limits convergence by introducing sequence biases, akin to online gradient descent with non-decaying step sizes.

Our theoretical results in Theorem \ref{lemma:multi layer} focus on multi-layer GLA without Multi-Layer Perceptron (MLP) layers to isolate and analyze the effects of the gating mechanism. However, MLP layers, a key component of standard Transformers, facilitate further nonlinear feature transformations and interactions, potentially enhancing GLA's expressive power. Future work could explore the theoretical foundations of integrating MLPs into GLA and analyze the optimization landscape of general gated attention models, aligning them more closely with conventional Transformer architectures \citep{gu2023mamba,dao2024transformers,peng2024eagle}.

\subsection{{GLA with scalar gating}}

Theorem~\ref{thm:gla=wpgd} establishes a connection between one-layer GLA (cf.~\eqref{eq:GLA}) and one-step WPGD (cf.~\eqref{eq:wpgd}), where the weighting in WPGD corresponds to the gating $G(\z_i)=\Gb_i$ in GLA, as detailed in Theorem~\ref{thm:gla=wpgd}. Now let us consider the widely used types of gating functions, such as $\Gb_i=\bal_i\onebb^\top_{d+1}$ 
\citep{yang2023gated,katsch2023gateloop,qin2024hgrn2,peng2024eagle} or $\Gb_i=\gamma_i\onebb_{(d+1),(d+1)}$ \citep{dao2024transformers,beck2024xlstm,peng2021random,sun2024you} where $\bal_i\in\R^{d+1}$ and $\gamma_i\in\R$. In both cases, the gating matrices in \eqref{weighting} take the form of $\begin{bmatrix}
    *&*\\g_i\onebb_d^\top&*
\end{bmatrix}$ for some $g_i\in\R$, thus simplifying the predictor to a sample-wise WPGD, as given by 
\begin{align}
f_{\gla}(\Z)=\hat\bt^\top\x,\quad\textnormal{with}\quad\hat\bt=\Pb\X^\top(\bom\odot\y),\label{wpgd y}
\end{align}
where $\Pb=\Pq\Pk^\top$ and $\bom=[g_{1:n+1}~\cdots~g_{n:n+1}]^\top\in\R^n$. In the remainder, we will mostly focus on the one-layer GLA with scalar gating as presented in \eqref{wpgd y}. 
\section{Optimization landscape of WPGD}\label{sec:optim gd}
In this section, we explore the problem of learning the optimal {sample-wise WPGD} algorithm described in \eqref{wpgd y}, a key step leading to our analysis of GLA. The problem is as follows. Recap from \eqref{PGD loss} that we are given the tuple $(\x, y, \X, \y) \sim \Dc$, where $\X \in \R^{n \times d}$ is the input matrix, $\y \in \R^n$ is the label vector, $\x \in \R^d$ is the query, and $y \in \R$ is its associated label. The goal is to use $\X, \y$ to predict $y$ given $\x$ via the one-step WPGD prediction $\hat{y} = \x^\top \hat{\bt}$, with $\hat{\bt}$ as in \eqref{wpgd y}. The algorithm learning problem is given by \eqref{eqn:wpgd:risk} which minimizes the WPGD risk $\E_{\Dc}[\left(y-\x^\top \Pb \X(\bom \odot \y)\right)^2]$.

Prior research \citep{mahankali2023one,li2024fine,ahn2024transformers} has studied the problem of learning PGD when input-label pairs follow an i.i.d. distribution. It is worth noting that while \cite{li2024fine} establishes a connection between H3-like models and \eqref{wpgd y} similar to ours, their work assumes that the optimal $\bom$ consists of all ones and does not specifically explore the optimization landscape of $\bom$ when in-context samples are not generated from the same task vector $\bt$.  
Departing from this, we introduce a realistic model where each input-label pair is allowed to come from a distinct task.

\begin{definition}[Correlated Task Model] \label{corr task def} Suppose $\bt_i\in\R^d\sim \Nn(0,\Iden)$ are jointly Gaussian for all $i\in[n+1]$.  Define 
\begin{align}\label{eqn:Randr}
\bt:=\bt_{n+1},\quad \B:=[\bt_1~\dots~\bt_n]^\top,\quad \Rb:=\frac{1}{d}\E[\B\B^\top],\quad \textnormal{and}\quad \rb:=\frac{1}{d}\E[\B\bt].    
\end{align}
\end{definition}
Note that in \eqref{eqn:Randr}, we have $\Rb \in \R^{n \times n}$  and $\rb \in \R^n$, with normalization ensuring that the entries of $\Rb$ and $\rb$ lie in the range $[-1,1]$, corresponding to correlation coefficients. Further, due to the joint Gaussian nature of the tasks, for any $i,j\in[n+1]$, the residual $\bt_i-r_{ij}\bt_j$ is independent of $\bt_j$.
\begin{definition}[Multitask Distribution]\label{def:multitask}
Let $(\bt_i)_{i=1}^{n+1}$ be drawn according to the correlated task model of Definition~\ref{corr task def}, $(\x_i)_{i=1}^{n+1}\in\R^d$ be i.i.d. with $\x_i\sim\Nn(0,\bSi)$, and $y_i\sim\Nn(\x_i^\top\bt_i,\sigma^2)$ for all $i\in[n+1]$.
\end{definition}
%
%
\begin{definition}
Let the eigen decompositions of $\bSi$ and $\Rb$ be denoted by 
$\bSi = \Ub \textnormal{diag}(\s) \Ub^\top$ and $\Rb = \Eb \textnormal{diag}(\bla) \Eb^\top$, 
where $\s = [s_1~ \cdots~ s_d]^\top \in \R^d_{++}$ and $\bla = [\lambda_1~\cdots~ \lambda_n]^\top \in \R_{+}^n$. Let $s_{\min}$ and $s_{\max}$ denote the smallest and largest eigenvalues of $\bSi$, respectively. Further, let $\lambda_{\min}$ and $\lambda_{\max}$ denote the \textit{nonzero} smallest and largest eigenvalues of $\Rb$. Define the effective spectral gap of $\bSi$ and $\Rb$, respectively, as
\begin{align}\label{eqn:spec:gap}
\Delta_{\bSi} := s_{\max}- s_{\min}, \quad\textnormal{and}  \quad \Delta_{\Rb} := \lambda_{\max}-\lambda_{\min}. 
\end{align}
\end{definition}

\begin{assumption}\label{assum:rb:in:eigenspace}
{The correlation vector $\rb$ from \eqref{eqn:Randr} lies in the range of $\Eb$}, i.e., $\rb = \Eb \ab$ for some nonzero  $\ab = [a_1~ \cdots~ a_n]^\top \in \R^n$. 
\end{assumption}

Assumption~\ref{assum:rb:in:eigenspace} essentially ensures that $\rb$ (representing the correlations between in-context tasks) can be expressed as a linear transformation of a vector \(\ab \) with at least one nonzero value. {Note that since $\rb$ lies in the range of the eigenvector matrix $\Eb$, it also lies in the range of $\Rb$ defined in \eqref{eqn:Randr}}. This guarantees that the correlation structure is non-degenerate, meaning that all elements of \( \rb \) are influenced by meaningful correlations. Assumption~\ref{assum:rb:in:eigenspace} avoids trivial cases where there are no correlations between tasks. By requiring at least one nonzero element in \( \ab \), the assumption ensures that the tasks are interrelated.

The following theorem characterizes the stationary points $(\Pb, \bom)$ of the WPGD objective in \eqref{eqn:wpgd:risk}.
\begin{theorem}\label{thm:station:wpgd}
Consider independent linear data as described in Definition~\ref{def:multitask}. Suppose Assumption~\ref{assum:rb:in:eigenspace} on the correlation vector $\rb$ holds. Let the functions $h_1: \mathbb{R}_{+} \to \mathbb{R}_{+}$ and $h_2: [1, \infty) \to \mathbb{R}_{+}$ be defined as
\begin{subequations}
\begin{align}
    h_1(\bar{\gamma}) &:= \left(\sum_{i=1}^n \frac{\lambda_i a_i^2}{ \left(1 + \lambda_i \bar{\gamma} \right)^2}\right) \left( \sum_{i=1}^n \frac{a_i^2}{\left(1 + \lambda_i \bar{\gamma} \right)^2} \right)^{-1}, \label{func:h} \\
    h_2(\gamma) &:= \left(1+ M\left(\sum_{i=1}^d \frac{s_i^2}{(M + s_i \gamma)^2}\right)\left(\sum_{i=1}^d \frac{s_i^3}{(M +  s_i \gamma )^2}\right)^{-1} \right)^{-1}, \label{func:g}
\end{align}
\end{subequations}
where $\{s_i\}_{i=1}^d$ and $\{\lambda_i\}_{i=1}^n$ are the eigenvalues of $\bSi$ and $\Rb$, respectively; $\{a_i\}_{i=1}^n$ are given in Assumption~\ref{assum:rb:in:eigenspace}; and $M = \sigma^2 + \sum_{i=1}^d s_i$.

The risk function $\Lc_{\wpgd}(\Pb, \bom)$ in \eqref{eqn:wpgd:risk} has a stationary point $(\Pb^\star, \bom^\star)$, up to rescaling, defined as
\begin{align}\label{eqn:pstar_and_wstar}
    \Pb^\star &= \bSi^{-\frac{1}{2}} \left( \frac{\gamma^\star }{M} \cdot \bSi + \Iden \right)^{-1} \bSi^{-\frac{1}{2}}, \quad \textnormal{and} \quad \bom^\star = \left( h_2(\gamma^\star) \cdot \Rb + \Iden \right)^{-1} \rb,
\end{align}
where $\gamma^\star$ is a fixed point of composite function \( h_1(h_2(\gamma))+1\).
\end{theorem}

Theorem~\ref{thm:station:wpgd} characterizes the stationary points $(\Pb^\star, \bom^\star)$, which exist up to re-scaling. This result presents the first landscape analysis of GLA for the joint learning of $(\Pb, \bom)$, while also exploring the stationary points $(\Pb^\star, \bom^\star)$.  In the following, we provide mild conditions on  effective spectral gaps of $\Rb$ and $\bSi$ under which a unique (global) minimum $(\Pb^\star, \bom^\star)$ exists.

\begin{theorem}[Uniqueness of the WPGD Predictor]\label{thm:unique:wpgd} 
Consider independent linear data as given in Definition \ref{def:multitask}. Suppose Assumption~\ref{assum:rb:in:eigenspace} on the correlation vector $\rb$ holds, and 
\begin{align} \label{eqn:gap:cond}
\Delta_{\bSi}\cdot \Delta_{\Rb} <M+s_{\min},    
\end{align}
where \( \Delta_{\bSi} \) and \( \Delta_{\Rb} \) denote the effective spectral gaps of \( \bSi \) and \( \Rb \), respectively, as given in  \eqref{eqn:spec:gap}; \( s_{\min} \) is the smallest eigenvalue of \( \bSi \); and  $M= \sigma^2 + \sum_{i=1}^d s_i$.
\begin{enumerate}[label={\textnormal{\textbf{T\arabic*.}}}, wide, labelindent=0pt, itemsep=0pt]
\item \label{item:fixedpoint} The composite function \( h_1(h_2(\gamma)) + 1\) is a contraction mapping and admits a unique fixed point \( \gamma = \gamma^\star \).
\item \label{item:globalmin} The function $\Lc_{\wpgd}(\Pb, \bom)$ has a unique (global) minima $(\Pb^\star, \bom^\star)$, up to re-scaling, given by \eqref{eqn:pstar_and_wstar}.
\end{enumerate}
\end{theorem}
%

\textit{Proof sketch}. Let \( \gamma := \frac{\bom^\top \Rb \bom}{\|\bom\|^2}+1 \). Note that \( \gamma \geq 1 \) since \( \Rb \) is positive semi-definite. From the first-order optimality condition, the solution to \eqref{eqn:wpgd:risk} takes the following form:
\begin{align}\label{eqn:pgamma_and_wgamma}
\Pb(\gamma) &= C(\rb, \bom, \bSi) \cdot \bSi^{-\frac{1}{2}} \left( \frac{\gamma}{M} \cdot \bSi + \Iden \right)^{-1}\bSi^{-\frac{1}{2}}, \quad\text{and}\quad\bom(\gamma) = c(\rb, \bom, \bSi) \cdot \Big(h_2(\gamma) \cdot \Rb + \Iden \Big)^{-1} \rb,    
\end{align}    
for some constants \( C(\rb, \bom, \bSi) \) and \( c(\rb, \bom, \bSi) \).

Substituting the expression for \( \bom(\gamma) \) into \( \gamma = \frac{\bom^\top \Rb \bom}{\|\bom\|^2}+1 \), and applying Assumption~\ref{assum:rb:in:eigenspace}, we obtain the equation \( \gamma = h_1(h_2(\gamma)) +1 \). We then show that whenever \( \Delta_{\bSi} \cdot \Delta_{\Rb} < M + s_{\min} \), the mapping \( h_1(h_2(\gamma)) +1\) is a contraction (see Lemma~\ref{lem:fixedpoint}). By the Banach Fixed-Point Theorem, this guarantees the existence of a unique fixed point \( \gamma = \gamma^\star \), where \( \gamma^\star = h_1(h_2(\gamma^\star)) +1 \). Finally, substituting \( \gamma^\star \) into \eqref{eqn:pgamma_and_wgamma} implies that \( (\Pb^\star, \bom^\star) \), as given in \eqref{eqn:pstar_and_wstar}, is a unique (global) minima of \eqref{eqn:wpgd:risk}, up to re-scaling. See Appendix~\ref{sec:app:wpgd} for the complete proof of Theorem~\ref{thm:unique:wpgd}. \hfill \qed

Theorem~\ref{thm:unique:wpgd} establishes mild conditions under which a unique (global) minimum $(\Pb^\star, \bom^\star)$ exists, up to scaling invariance, and guarantees the uniqueness of the associated WPGD algorithm. It provides the first global landscape analysis for GLA and generalizes prior work \citep{li2024fine, ahn2024transformers} on the global landscape by extending the optimization properties of linear attention to the more complex \textit{nonconvex} GLA with joint $(\Pb, \bom)$ optimization. 

\begin{enumerate}[label={\textnormal{\textbf{Remark \arabic*.}}}, wide, labelindent=0pt, itemsep=0pt]
\item \label{item:rem2} 
An interesting observation about the optimal gating parameter $\bom^\star$ is its connection to the correlation matrix $\Rb$, which captures the task correlations in a multitask learning setting. Specifically, the optimal gating given in \eqref{eqn:pstar_and_wstar} highlights how $\bom^\star$ depends directly on both the task correlation matrix $\Rb$ and the vector $\rb$, which encodes the correlations between the tasks and the target task. 
\item \label{item:rem3} Condition~\eqref{eqn:gap:cond} provides a \textit{sufficient} condition for the uniqueness of a fixed point. This implies that whenever $\Delta_{\bSi} \cdot \Delta_{\Rb} < M + s_{\min}$, the mapping $h_1(h_2(\gamma))+1$ is a contraction, ensuring the existence of a unique fixed point. However, there may be cases where the mapping $h_1(h_2(\gamma))+1$ does not satisfy Condition~\eqref{eqn:gap:cond}, yet a unique fixed point (and a unique global minimum) still exists. This is because the Banach Fixed-Point Theorem does not provide a \textit{necessary} condition.
\end{enumerate}

\begin{corollary}\label{coroll:unique}
Suppose $\bSi = \Iden$. Then, $\Delta_{\bSi} = 0$, satisfying Condition~\eqref{eqn:gap:cond}, and we have $h_2(\gamma^\star) = \frac{1}{d + \sigma^2 + 1}$, which yields
\begin{equation}\label{eqn:opt:obj}
\Pb^\star = \Iden, \quad \textnormal{and} \quad 
\bom^\star = \left( \Rb + (d + \sigma^2 + 1)\Iden \right)^{-1} \rb.
\end{equation}
Thus, the optimal risk $\Lc^\st_\wpgd$ defined in \eqref{eqn:wpgd:risk} is given by
\begin{equation}
    \Lc_\wpgd^
    \st= d+\sigma^2  -  d \cdot \rb^\top\left( \Rb + (d + \sigma^2 + 1)\Iden \right)^{-1} \rb.\label{optim wpgd risk}
\end{equation}
\end{corollary}


\section{Optimization landscape of GLA}\label{gla sec}
In Section~\ref{sec plain wpgd}, we demonstrated that GLA implements a data-dependent WPGD algorithm. Building on this, in Section~\ref{sec:optim gd}, we analyze the optimization landscape for minimizing the one-step WPGD risk (cf.~\eqref{eqn:wpgd:risk}) and show that a unique solution (up to rescaling) achieves the global minimum of the WPGD algorithm. However, in GLA, the search space for $\bom$ is restricted and data-dependent, meaning that $\Lc_\wpgd^\st$ in \eqref{eqn:wpgd:risk} represents the best possible risk a GLA model can achieve. In this section, we analyze the loss landscape for training a one-layer GLA model and explore the scenarios under which GLA can reach the optimal WPGD risk.
\subsection{Multi-task prompt model}\label{sec:multi p}
Following Definitions~\ref{corr task def} and \ref{def:multitask}, we consider the multi-task prompt setting with \( K \) correlated tasks \((\bt_k)_{k=1}^K\) and one query task \(\bt\) with distribution $\Nc(0,\Iden)$. For each correlated task $k\in[K]$, a prompt of length \( n_k \) is drawn, consisting of IID input-label pairs \(\{(\x_{ik}, y_{ik})\}_{i=1}^{n_k}\) where $y_{ik}\sim\Nc(\bt_k^\top\x_{ik},\sigma^2)$ to obtain sequence $\Z_k=\begin{bmatrix}
    \x_{1k}&\cdots&\x_{n_k,k}\\y_{1k}&\cdots&y_{n_k,k}
\end{bmatrix}^\top$. Let \( n := \sum_{k=1}^K n_k \). The query is given by \(\z_{n+1} = [\x_{n+1}~~0]^\top\) with the corresponding label \( y_{n+1} \sim \Nc(\x_{n+1}^\top\bt, \sigma^2) \). All the features $\x_{ik}$ for $i\in[n_k],k\in[K]$ and query feature $\x_{n+1}$ are IID sampled from $\Nc(\zerobb,\bSi)$.

These sequences \((\Z_k)_{k=1}^K\), along with the query token \(\z_{n+1}\), are concatenated to form a single prompt \(\Z\). Recall \eqref{eq:GLA}, and let \( f_{\gla}(\Z) \) denote the GLA prediction as defined in Theorem~\ref{thm:gla=wpgd}. Additionally, consider the model construction presented in \eqref{att mtx res}, where \(\Pq, \Pk \in \R^{d \times d}\) are trainable parameters. The GLA optimization problem is described as follows:
\begin{equation}\label{eqn:gla:risk}
\begin{split}
&\Lc_\gla^\st := \min_{\Pk, \Pq\in\R^{d\times d}, G\in\Gc}~\Lc_\gla(\Pk, \Pq, G), \quad \\
&\text{where} \quad \Lc_\gla(\Pk, \Pq, G) = \E\left[\left(y - f_{\gla}(\Z)\right)^2\right]. 
\end{split}
\end{equation}
Here, $G(\cdot)$ represents the gating function and $\Gc$ denotes the function search space, which is determined by the chosen gating strategies (cf. Table 1 in \cite{yang2023gated}). 

\begin{figure}[!t]
\centering
\begin{subfigure}{0.24\linewidth}    
\centering
\includegraphics[width=\linewidth]{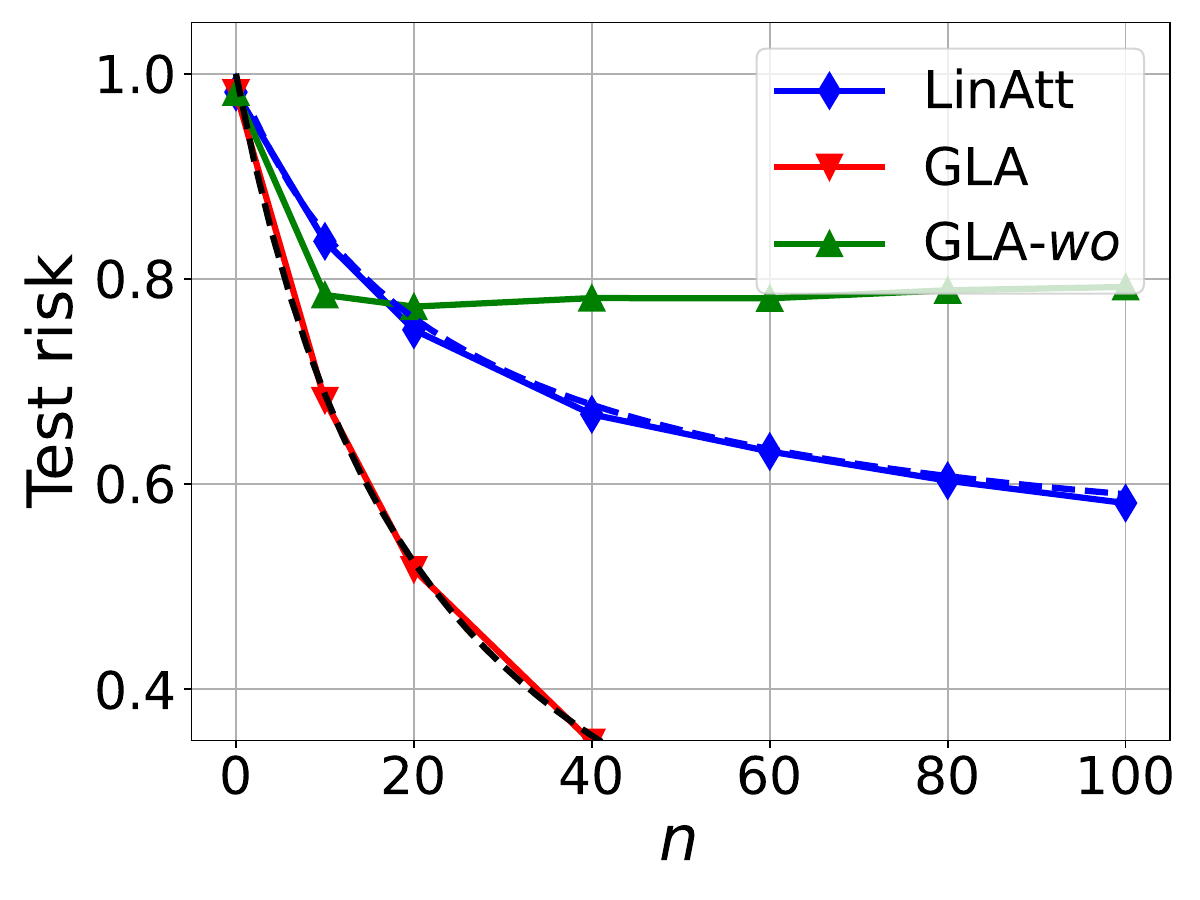}
\caption{$(r_1,r_2)=(0,1)$}\label{fig:0 1}
\end{subfigure}
\begin{subfigure}{0.24\linewidth}    
    \centering
    \includegraphics[width=\columnwidth]{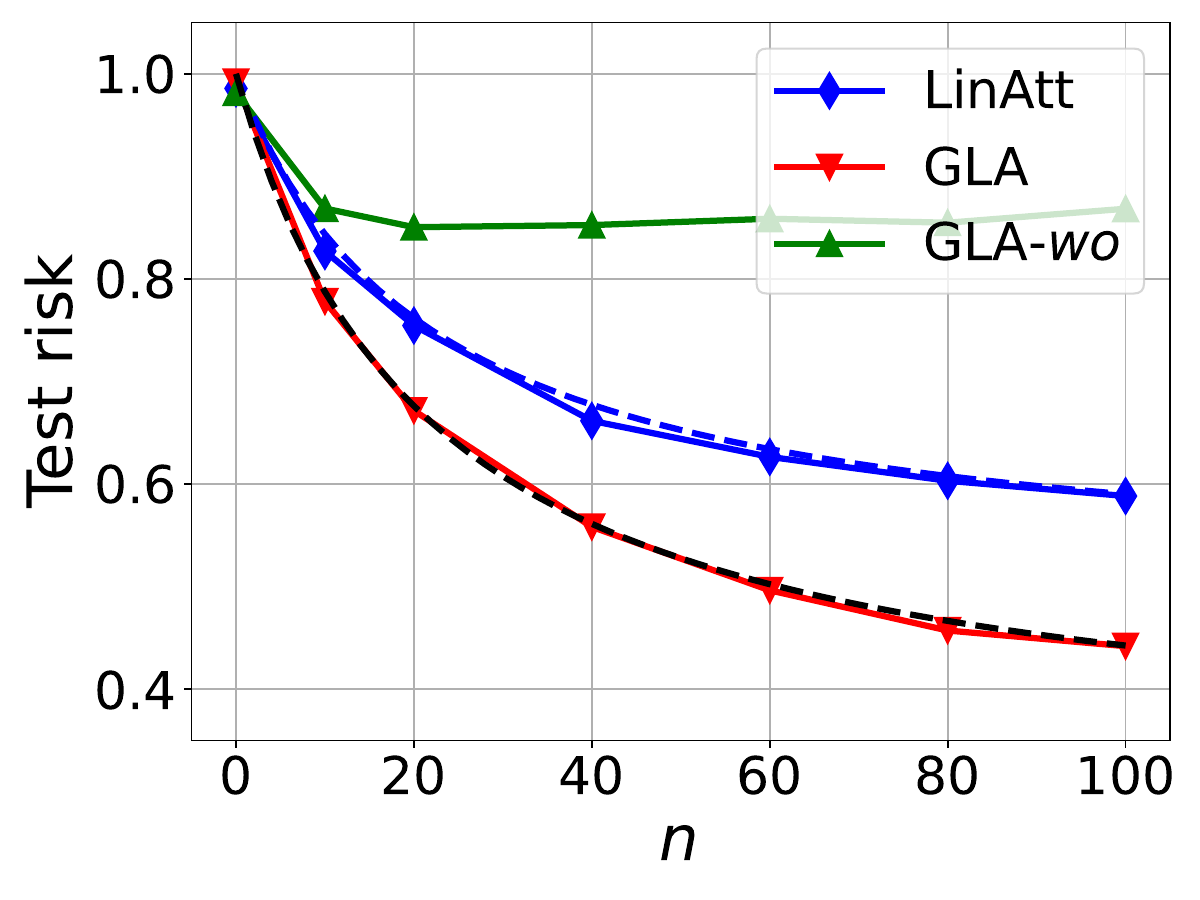}
\caption{$(r_1,r_2)=(0.2,0.8)$}\label{fig:0.2 0.8}
\end{subfigure}
\begin{subfigure}{0.24\linewidth}    
\centering
\includegraphics[width=\columnwidth]{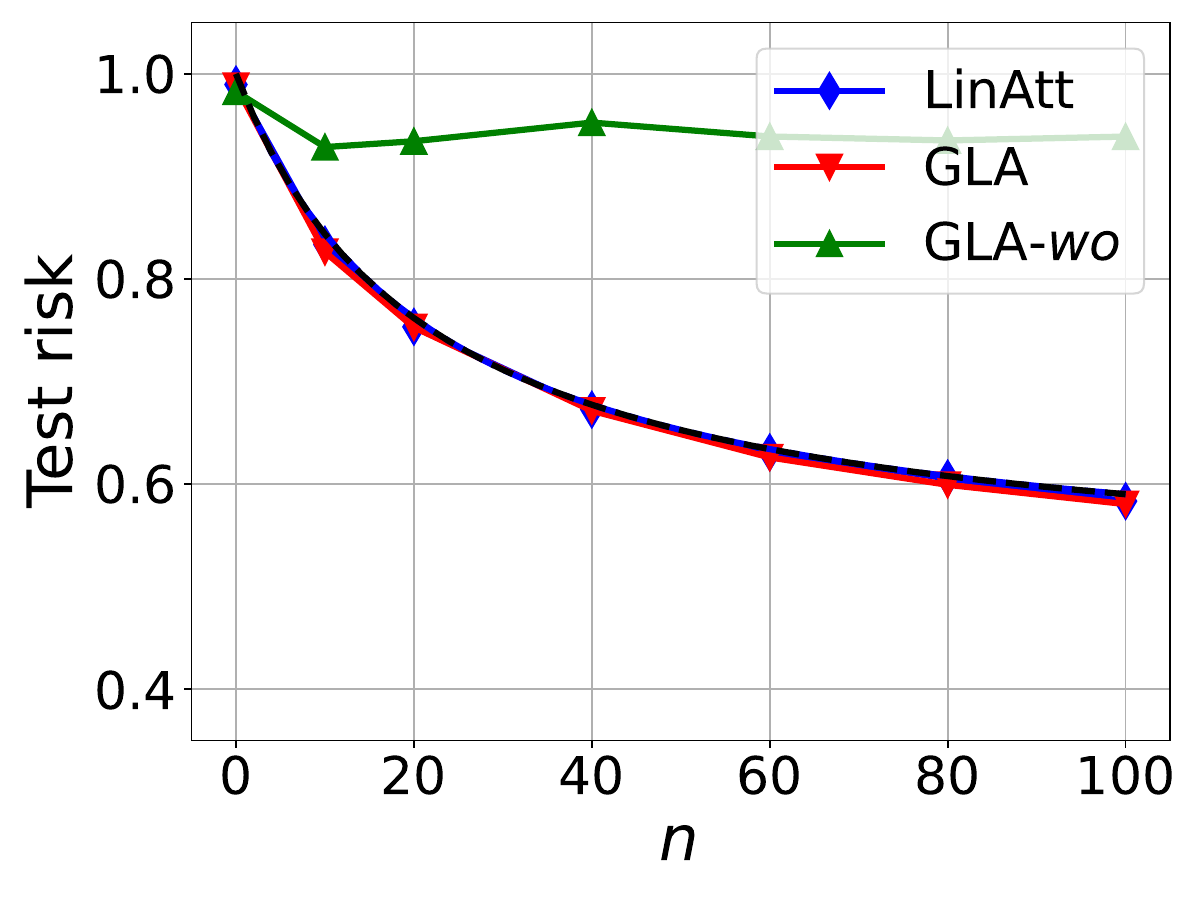}
\caption{$(r_1,r_2)=(0.5,0.5)$}\label{fig:0.5 0.5}
\end{subfigure}
\begin{subfigure}{0.24\linewidth}    
\centering
\includegraphics[width=\columnwidth]{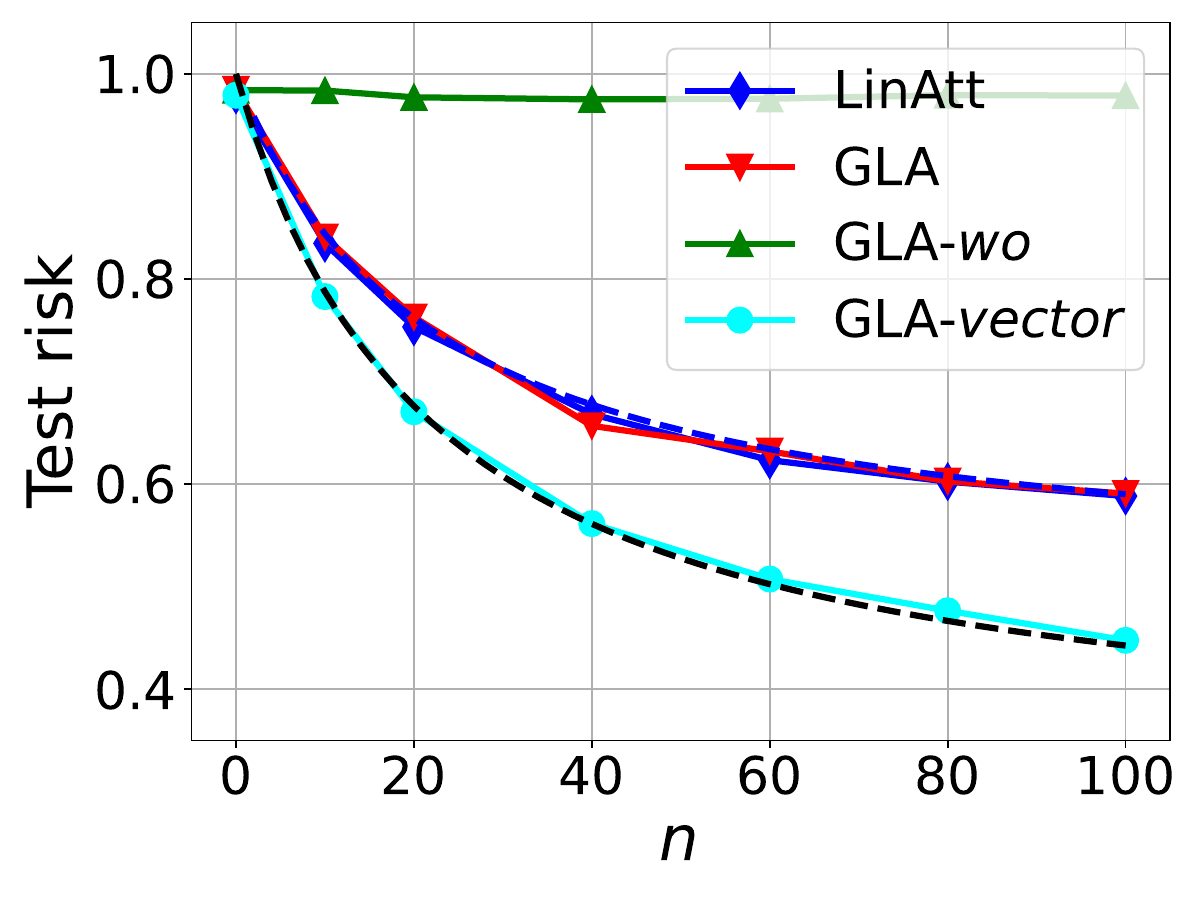}
\caption{$(r_1,r_2)=(0.8,0.2)$}\label{fig:0.8 0.2}
\end{subfigure}
\caption{\small{We consider four different types of model training: \texttt{LinAtt} (blue solid): Standard linear attention training. \texttt{GLA} (red solid): GLA training using prompts with delimiters (see \eqref{mtl prompt}) and scalar gating. \texttt{GLA-\emph{wo}} (green solid): GLA training using prompts without delimiters and with scalar gating. \texttt{GLA-\emph{vector}} (cyan solid): GLA training using prompts with delimiters and vector gating. The blue and black dashed curves represent the optimal linear attention and WPGD risks from \eqref{optim att risk} and \eqref{optim wpgd risk}, respectively, as the number of in-context examples $n$ increases. Implementation details are provided in Appendix~\ref{sec:exp}.}}\label{fig:k=2}
\vspace{-10pt}
\end{figure}

Note that 1) the task vectors $(\bt_k)_{k=1}^K$ are not explicitly shown in the prompt,
2) in-context examples $(\x_{ik}, y_{ik})$ are randomly drawn, and 3) the gating function is applied to the tokens/input samples $(\Z_k)_{k=1}^K$. Given the above three evidences, the implicit weighting induced by the GLA model varies across different prompts, and it prevents the GLA from learning the optimal weighting. 
To address this, we introduce delimiters to mark the boundary of each task. 
Let $(\db_k)_{k=1}^{K}$ be the delimiters that determine stop of the tasks.
Specifically, the final prompt is given by 
\begin{align}
\Z=\begin{bmatrix}\Z_1^\top&\db_1&\cdots&\Z_K^\top&\db_K&\z_{n+1}\end{bmatrix}^\top.\label{mtl prompt}
\end{align}
Additionally, to decouple the influence of gating and data, we envision that each token is $\z_i=[\x_i^\top~~y_i~~\cb_i^\top]^\top$ where $\cb_i\neq\zerobb\in\R^p$ is the contextual features with $p$ being its dimension and $(\x_i,y_i)$ are the data features. 
Let $\cbb_0,\cdots,\cbb_K\in\R^p$ be $K+1$ contextual feature vectors.
\begin{itemize}
    \item For task prompts $\Z_k$: Contextual features are set to a fixed vector $\cbb_0\neq \zerobb$.
    \item For delimiters $\db_k$: Data features are set to zero (e.g., $\x_i=\zerobb$ and $y_i=0$) so that $\db_k=[\zerobb_{d+1}^\top~\cbb_k^\top]^\top$.
\end{itemize}
Explicit delimiters have been utilized in addressing real-world problems \citep{wang2024one,asai2022attempt,dun2023sweeping} due to their ability to improve efficiency and enhance generalization, particularly in task-mixture or multi-document scenarios. To further motivate the introduction of \((\db_k)_{k=1}^K\), we present in Figure~\ref{fig:k=2} the results of GLA training with and without delimiters, represented by the red and green curves, respectively. The black dashed curves indicate the optimal WPGD loss, \(\Lc^\st_\wpgd\), under different scenarios. Notably, training GLA without delimiters (the green solid curve) performs strictly worse. In contrast, training with delimiters can achieve optimal performance under certain conditions (see Figures~\ref{fig:0 1}, \ref{fig:0.2 0.8}, and \ref{fig:0.5 0.5}). Theorem~\ref{thm:gla optim}, presented in the next section, provides a theoretical explanation for these observations, including the misalignment observed in Figure~\ref{fig:0.8 0.2}. Further discussion and experimental details are provided in Section~\ref{sec:gla optim} and Appendix~\ref{sec:exp}.

\subsection{Loss landscape of one-layer GLA}\label{sec:gla optim}
Given the input tokens with extended dimension, to ensure that GLA still implements WPGD as in Theorem~\ref{thm:gla=wpgd}, we propose the following model construction. 
\begin{align}\Wt_k=\begin{bmatrix}
    \Wk&\zerobb\\
    \zerobb&\zerobb
\end{bmatrix},\quad\Wt_q=\begin{bmatrix}
    \Wq&\zerobb\\\zerobb&\zerobb
\end{bmatrix},\quad\text{and}\quad\tilde\W_v=\begin{bmatrix}\Wv&\zerobb\\\zerobb&\zerobb\end{bmatrix}.\label{att mtx res delimiter}
\end{align}
Here, $\Wt_{k,q,v}\in\R^{(d+p+1)\times(d+p+1)}$ and $\W_{k,q,v}\in\R^{(d+1)\times(d+1)}$ are constructed via \eqref{att mtx res}. 
The main idea is to set the last $p$ rows and columns of attention matrices to zeros, ensuring that the delimiters do not affect the final prediction. 
\begin{assumption}\label{assum delimiter gate}
Contextual feature vectors $\cbb_0,\cdots,\cbb_K$ are linearly independent, and 
    activation function $\phi(z):\R\to[0,1]$ is continuous, satisfying $\phi(-\infty) = 0$ and $\phi(+\infty) = 1$.
\end{assumption}
\begin{assumption}\label{assum order}
    The correlation between context tasks $(\bt_k)_{k=1}^K$ and query task $\bt$ satisfies $\E[\bt_i^\top\bt_j]=0$ and $\E[\bt_i^\top\bt]\leq\E[\bt_j^\top\bt]$ for  all $1\leq i\leq j\leq K$.
\end{assumption}

Given context examples $\{(\X_k,\y_k):=(\x_{ik},y_{ik})_{i=1}^{n_k}\}_{k=1}^{K}$, define the concatenated data $(\X,\y)$ as follows:
\begin{align}
\X=\begin{bmatrix}
    \X_1^\top&\cdots&\X_K^\top
\end{bmatrix}^\top\in\R^{n\times d}, \quad\text{and}\quad\y=\begin{bmatrix}
    \y_1^\top&\cdots&\y_K^\top
\end{bmatrix}^\top\in\R^{n}.\label{K task X y}
\end{align}
Based on the assumptions above, we are able to establish the equivalence between optimizing one-layer GLA and optimizing one-step WPGD predictor {under scalar gating}.
\begin{theorem} [{Scalar Gating}]\label{thm:gla optim}
Suppose Assumption~\ref{assum delimiter gate} holds. Recap the function $\Lc_\wpgd(\Pb,\bom)$ from \eqref{eqn:wpgd:risk} with dataset $(\X,\y)$ defined in \eqref{K task X y}. Consider \eqref{eq:GLA} with input prompt $\Z$ defined in \eqref{mtl prompt}, the model constructions described in \eqref{att mtx res delimiter}, and  scalar gating $G(\z)=\phi(\w_g^\top\z)\onebb_{(d+p+1)\times(d+p+1)}$, where $\w_g$ is a trainable parameter. Then, the optimal risk $\Lc^\st_\gla$ defined in \eqref{eqn:gla:risk} obeys
    \begin{align}
{\Lc_\gla^\st=\Lc_{\wpgd}^{\st,\Wc}, \quad\text{where}\quad\Lc_{\wpgd}^{\st,\Wc}:=\min_{\Pb\in\R^{d\times d},\bom\in\Wc}~\Lc_\wpgd(\Pb,\bom).}\label{gla=wpgd W}
    \end{align}
    Here, 
    $
    \Wc:=\left\{\left[\omega_1\onebb_{n_1}^\top~\cdots~\omega_K\onebb_{n_K}^\top\right]^\top\in\R^{n}~\Big|~0\leq\omega_i\leq\omega_j\leq1,~\forall 1\leq i\leq j\leq K\right\}.
    $
    
    Additionally, suppose Assumption~\ref{assum order} holds and $n_i=n_j$, for all  $i,j\in[K]$. Let $\Lc_\wpgd^\st$ be the optimal WPGD risk (cf.~\eqref{eqn:wpgd:risk}). Then  $\Lc^\st_\gla$ satisfies
    \begin{align}
    \Lc_\gla^\st=\Lc_\wpgd^\st.\label{gla=wpgd}
    \end{align}
\end{theorem}
Assumption~\ref{assum delimiter gate} ensures that any $\bom$ in $\Wc$ can be achieved by an appropriate choice of gating parameters. Furthermore, Assumption~\ref{assum order} guarantees that the optimal choice of $\bom$ under the WPGD objective lies within the search space $\Wc$. The proof is provided in Appendix~\ref{app:proof gla optim}.

In Figure~\ref{fig:k=2}, we conduct model training to validate our findings. Consider the setting where $K=2$ and let $(r_1, r_2) = \left(\E[\bt_1^\top\bt]/d, \E[\bt_2^\top\bt]/d\right)$. In Figures~\ref{fig:0 1}, \ref{fig:0.2 0.8}, and \ref{fig:0.5 0.5}, Assumption~\ref{assum order} holds, and the GLA results (shown in solid red) align with the optimal WPGD risk (represented by the dashed black curves), validating \eqref{gla=wpgd}. However, in Figure~\ref{fig:0.8 0.2}, since $r_1 > r_2$, Assumption~\ref{assum order} does not hold, and as a result, the optimal GLA loss $\Lc^\st_\gla$ obtained from \eqref{gla=wpgd W} is lower than the optimal WPGD loss $\Lc^\st_\wpgd$. Further experimental details are deferred to Appendix~\ref{sec:exp}.

\paragraph*{Loss landscape of vector gating.} Till now, much of our discussion has focused on the scalar gating setting.
It is important to highlight that, even in the scalar-weighting context, analyzing the WPGD problem remains non-trivial due to the joint optimization over $(\Pb, \bom)$. However, as demonstrated in Theorem~\ref{thm:gla optim}, scalar gating can only express weightings within the set $\Wc$. If Assumption~\ref{assum order} does not hold, $\Lc^\st_\gla$ cannot achieve the optimal WPGD loss (see the misalignment between red solid curve, presenting $\Lc^\st_\gla$, and black dashed curve, presenting $\Lc^\st_\wpgd$ in Figure~\ref{fig:0.8 0.2}). We argue that vector gating overcomes this limitation by applying distinct weighting mechanisms across different dimensions, facilitating stronger expressivity. 
\begin{theorem} [{Vector Gating}] \label{thm:gla optim vector}
 Suppose Assumption~\ref{assum delimiter gate} holds.  
 Consider \eqref{eq:GLA} with input prompt $\Z$ from \eqref{mtl prompt}, the model constructions from \eqref{att mtx res delimiter} but with $\tilde\W_v=[\zerobb_{(d+p+1)\times d}~~\ub~~\zerobb_{(d+p+1)\times p}]^\top$, and a vector gating $G(\z)=\phi(\W_g\z)\onebb^\top_{d+p+1}$. Recap Problem \eqref{eqn:gla:risk} with prediction defined as  $f_{\gla}(\Z):=\ob_{n+1}^\top\hb$, where $\hb\in\R^{d+p+1}$ is a linear prediction head. Here, $\ub$, $\W_g$ and $\hb$ are trainable parameters. Let $\Lc_{\gla\text{-}v}^\st$ denote its optimal risk, and $\Lc_{\wpgd}^\st$ be defined as in \eqref{eqn:wpgd:risk}.
    Then, the optimal risk obeys $
    \Lc_{\gla\text{-}v}^\st=\Lc_\wpgd^\st$.
\end{theorem}
{In Theorem~\ref{thm:gla optim}, the equivalence between $\Lc_{\gla}^\st$ and $\Lc_{\wpgd}^\st$ is established only when both Assumptions~\ref{assum delimiter gate} and \ref{assum order} are satisfied. In contrast, Theorem~\ref{thm:gla optim vector} demonstrates that applying vector gating requires only Assumption~\ref{assum delimiter gate} to establish $\Lc_{\gla\text{-}v}^\st = \Lc_{\wpgd}^\st$. Specifically, under the bounded activation model of Assumption~\ref{assum delimiter gate}, scalar gating is unable to express non-monotonic weighting schemes. For instance, suppose there are two tasks: Even if Task 1 is more relevant to the query, Assumption~\ref{assum delimiter gate} will assign a higher (or equal) weight to examples in Task 2 resulting in sub-optimal prediction. Theorem \ref{thm:gla optim vector} shows that vector gating can avoid such bottlenecks by potentially encoding tasks in distinct subspaces. 
To verify these intuitions}, in Figure~\ref{fig:0.8 0.2}, we train a GLA model with vector gating 
and results are presented in cyan curve, which outperform the scalar gating results (red solid) and align with the optimal WPGD loss (black dashed). 

\paragraph*{Loss landscape of One-layer linear attention.} Given the fact that linear attention is equivalent to \eqref{eq:GLA} when implemented with all ones gating, that is, $\Gb_i\equiv\onebb$, we derive the following corollary. Consider training a standard single-layer linear attention, i.e.,  by setting $\Gb_i\equiv\onebb$ in \eqref{eq:GLA}, and let $f_{\att}(\Z)$ be its prediction. Let  $\Lc_\att^\st$ be the corresponding optimal risk following \eqref{eqn:gla:risk}.  
\begin{corollary}\label{corol:att}
Consider a single-layer linear attention following model construction in \eqref{att mtx res} and let linear data as given in Definition~\ref{def:multitask}. Let $\Rb$ and $\rb$ be the corresponding correlation matrix and vector as defined in Definition~\ref{corr task def}. 
Suppose $\bSi=\Iden$. Then, the optimal risk obeys
\begin{align}
\Lc^\st_\att:=\min_{\Pb\in\R^{d\times d}}\Lc_\wpgd(\Pb,\bom=\onebb)=d+\sigma^2 - \frac{d(\onebb^\top \rb)^2}{n(d+\sigma^2+1) + \onebb^\top \Rb \onebb}.\label{optim att risk}
\end{align}
\end{corollary}
{
\begin{corollary}[Benefit of Gating]\label{corol:att vs gla}
Consider the same setting as discussed in Corollary~\ref{corol:att}, and suppose Assumption~\ref{assum delimiter gate} holds. Then, we have that $\Lc^\st_\att \geq \Lc^\st_\gla$. Additionally, if Assumption~\ref{assum order} holds, we obtain
\[
\Lc^\st_\att - \Lc^\st_\gla = d \cdot \rb^\top\left(\Rb_+^{-1} - \frac{\onebb\onebb^\top}{\onebb^\top \Rb_+ \onebb}\right)\rb \geq 0, \quad \textnormal{where}\quad   \Rb_+ := \Rb + \left(d + \sigma^2 + 1\right)\Iden.
\]
\end{corollary}
The proof of this corollary is directly from \eqref{optim wpgd risk}, \eqref{gla=wpgd} and \eqref{optim att risk}.}
In the Figure~\ref{fig:k=2}, blue solid curves represent the linear attention results
 and blue dashed are the theory curves following \eqref{optim att risk}. The two curves are aligned in all the subfigures, which validate our Corollary~\ref{corol:att}. More implementation details are deferred to Appendix~\ref{sec:exp}.

\section{Related work}

\paragraph*{Efficient sequence models.} Recent sequence model proposals -- such as RetNet \citep{sun2023retentive}, Mamba \citep{gu2023mamba}, xLSTM~\citep{beck2024xlstm}, GLA Transformer~\citep{yang2023gated}, RWKV-6~\citep{peng2024eagle} -- admit efficient recurrent forms while being increasingly competitive with the transformer architecture with softmax-attention. However, we have a rather limited theoretical understanding of these architectures, especially, when it comes to their optimization landscape and ICL capabilities. \cite{park2024can,grazzi2024mamba} demonstrate that Mamba is effective and competitive with a transformer of similar size in various ICL tasks, whereas \cite{arora2024simple,jelassi2024repeat} establish theoretical and empirical shortcomings of recurrent models for solving recall tasks. It is worth mentioning that, GLA models also connect to state-space models and linear RNNs \citep{de2024griffin,orvieto2023resurrecting,gu2021efficiently,fu2022hungry}, as they could be viewed as time-varying SSMs \citep{dao2024transformers,sieber2024understanding}. {Finally, GLA models are also closely related to implicit self-attention frameworks. For example, the work by \cite{zimerman2024unified} on unified implicit attention highlights how models such as Mamba \citep{gu2023mamba} and RWKV~\citep{peng2023rwkv} can be viewed under a shared attention mechanism. Additionally, \cite{zong2024stock} leverage gated cross-attention for robust multimodal fusion, demonstrating another practical application of gated mechanisms. Both approaches align with GLA’s data-dependent gating, suggesting its potential for explainability and stable fusion tasks.}

\paragraph*{Theory of in-context learning.} The theoretical aspects of ICL has been studied by a growing body of works during the past few years  \citep{xieexplanation,von2023uncovering,gatmirycan,li2023transformers,collins2024context,wu2023many,fu2023transformers,lin2024dual,akyrek2023what,zhang2023trained}. A subset of these follow the setting of \citet{garg2022can} which investigates the ICL ability of transformers by focusing on prompts where each example is labeled by a task function from a specific function class, such as linear models. 
\citet{akyrek2023what} focuses on linear regression and provide a transformer construction that can perform a single step of GD based on in-context examples. Similarly, \citet{von2023transformers} provide a construction of weights in linear attention-only transformers that can replicate GD steps for a linear regression task on in-context examples. Notably, they observe similarities between their constructed networks and those resulting from training on ICL prompts for linear regression tasks. Building on these, \citet{zhang2024trained, mahankali2023one, ahn2024transformers} focus on the loss landscape of ICL for linear attention models. For a single-layer model trained on in-context prompts for random linear regression tasks, \citet{mahankali2023one, ahn2024transformers} show that the resulting model performs a single  preconditioned GD step on in-context examples in a test prompt, aligning with the findings of \citet{von2023transformers}.{  More recent work \citep{ding2023causallm} analyzes the challenges of causal masking in causal language models (causalLM), showing that their suboptimal convergence dynamics closely resemble those of online gradient descent with non-decaying step sizes. } Additionally, \cite{li2024fine} analyzes the landscape of the H3 architecture, an SSM, under the same dataset model. They show that H3 can implement WPGD thanks to its convolutional/SSM filter. However, their WPGD theory is restricted to the trivial setting of equal weights, relying on the standard prompt model with IID examples and shared tasks. In contrast, we propose novel multitask datasets and prompt models where nontrivial weighting is provably optimal. This allows us to characterize the loss landscape of WPGD and explore advanced GLA models, linking them to data-dependent WPGD algorithms.

\paragraph*{Optimization Landscape of Attention Mechanisms.}
The optimization behavior of attention mechanisms has become a rapidly growing area of research \citep{deora2023optimization, huang2023context, tian2023joma, fu2023transformers, li2024mechanics, ataee2024max, tarzanagh2023transformers, deng2023unmasking, makkuva2024attention, jeon2024information, zheng2023learn, collins2024context, chen2024provably, li2023theoretical, ildiz2024self, vasudeva2024simplicity, bao2024self, chen2024training, huang2024non, wang2024transformers, sun2025context}.
Particularly relevant to our work are studies investigating convex relaxation approaches to optimize attention models \citep{sahiner2022unraveling, ergen2022convexifying}, gradient-based optimization methods in vision transformers \citep{jelassi2022vision}, optimization dynamics in prompt-attention \citep{oymak2023role,ataee2024max}, implicit bias of gradient descent for attention  optimization \citep{ataee2024max, tarzanagh2023transformers, julistiono2024optimizing, li2024mechanics, thrampoulidis2024implicit, zhao2024implicit, vasudeva2024implicit, sheen2024implicit}, and optimization geometry analysis  of next-token prediction \citep{zhao2025geometry}. While these works mainly focus on standard attention, our work provides the first optimization landscape analysis of \textit{gated} attention, establishing conditions for a unique global minimum and providing comprehensive insights into the implications of data-dependent weighting in the optimization dynamics encountered during training.

\section{Discussion}




Our analysis is currently limited to GLA models under specific weight construction assumptions as outlined in Section~\ref{sec plain wpgd}, which may not capture the full expressivity of practical GLA implementations. The theoretical framework we developed requires certain structural constraints on the attention matrices (as specified in Equation~\ref{att mtx res}) to establish the connection with WPGD algorithms.

Several important directions for future research include:

\begin{enumerate}
    \item Extending our analysis to more general GLA architectures without the weight construction constraints, which would better reflect deployed models such as  Mamba \citep{gu2023mamba,dao2024transformers} and RWKV \citep{peng2024eagle}. 
    \item Investigating when delimiters are necessary for effective learning in multi-task settings. Our experiments show that without delimiters, GLA models perform suboptimally, but a deeper theoretical characterization of this phenomenon is needed.
    
    \item Exploring the GLA landscape where gating functions depend on input features in more complex ways than our current formulation allows.
    
   \item Analyzing the effects of incorporating MLP layers between GLA layers, which could enhance the model's expressive power beyond what we have characterized.
    
\end{enumerate}
Addressing these limitations would further bridge the gap between our theoretical understanding and the practical performance of state-of-the-art GLA-based sequence models.

\section*{Acknowledgements}

Y.L.~and S.O.~were supported in part by the NSF grants CCF2046816, CCF-2403075, CCF-2008020, the Office of Naval Research grant N000142412289, and by gifts/awards from Open Philanthropy, OpenAI, Amazon Research, and Google Research.
M.F.\ was supported in part by awards NSF TRIPODS II 2023166, CCF 2007036, CCF 2212261,
and CCF 2312775. 

\bibliography{refs,mainrefb}

\begin{thebibliography}{75}
\providecommand{\natexlab}[1]{#1}
\providecommand{\url}[1]{\texttt{#1}}
\expandafter\ifx\csname urlstyle\endcsname\relax
  \providecommand{\doi}[1]{doi: #1}\else
  \providecommand{\doi}{doi: \begingroup \urlstyle{rm}\Url}\fi

\bibitem[Ahn et~al.(2024)Ahn, Cheng, Daneshmand, and Sra]{ahn2024transformers}
Kwangjun Ahn, Xiang Cheng, Hadi Daneshmand, and Suvrit Sra.
\newblock Transformers learn to implement preconditioned gradient descent for
  in-context learning.
\newblock \emph{Advances in Neural Information Processing Systems}, 36, 2024.

\bibitem[Aky{\"u}rek et~al.(2023)Aky{\"u}rek, Schuurmans, Andreas, Ma, and
  Zhou]{akyrek2023what}
Ekin Aky{\"u}rek, Dale Schuurmans, Jacob Andreas, Tengyu Ma, and Denny Zhou.
\newblock What learning algorithm is in-context learning? investigations with
  linear models.
\newblock In \emph{The Eleventh International Conference on Learning
  Representations}, 2023.

\bibitem[Arora et~al.(2024)Arora, Eyuboglu, Zhang, Timalsina, Alberti, Zinsley,
  Zou, Rudra, and R{\'e}]{arora2024simple}
Simran Arora, Sabri Eyuboglu, Michael Zhang, Aman Timalsina, Silas Alberti,
  Dylan Zinsley, James Zou, Atri Rudra, and Christopher R{\'e}.
\newblock Simple linear attention language models balance the recall-throughput
  tradeoff.
\newblock \emph{arXiv preprint arXiv:2402.18668}, 2024.

\bibitem[Asai et~al.(2022)Asai, Salehi, Peters, and
  Hajishirzi]{asai2022attempt}
Akari Asai, Mohammadreza Salehi, Matthew~E Peters, and Hannaneh Hajishirzi.
\newblock Attempt: Parameter-efficient multi-task tuning via attentional
  mixtures of soft prompts.
\newblock \emph{arXiv preprint arXiv:2205.11961}, 2022.

\bibitem[Bao et~al.(2024)Bao, Hataya, and Karakida]{bao2024self}
Han Bao, Ryuichiro Hataya, and Ryo Karakida.
\newblock Self-attention networks localize when qk-eigenspectrum concentrates.
\newblock \emph{arXiv preprint arXiv:2402.02098}, 2024.

\bibitem[Beck et~al.(2024)Beck, P{\"o}ppel, Spanring, Auer, Prudnikova, Kopp,
  Klambauer, Brandstetter, and Hochreiter]{beck2024xlstm}
Maximilian Beck, Korbinian P{\"o}ppel, Markus Spanring, Andreas Auer,
  Oleksandra Prudnikova, Michael Kopp, G{\"u}nter Klambauer, Johannes
  Brandstetter, and Sepp Hochreiter.
\newblock {xLSTM}: Extended long short-term memory.
\newblock \emph{arXiv preprint arXiv:2405.04517}, 2024.

\bibitem[Brown(2020)]{brown2020language}
Tom~B Brown.
\newblock Language models are few-shot learners.
\newblock \emph{arXiv preprint arXiv:2005.14165}, 2020.

\bibitem[Chen \& Li(2024)Chen and Li]{chen2024provably}
Sitan Chen and Yuanzhi Li.
\newblock Provably learning a multi-head attention layer.
\newblock \emph{arXiv preprint arXiv:2402.04084}, 2024.

\bibitem[Chen et~al.(2024)Chen, Sheen, Wang, and Yang]{chen2024training}
Siyu Chen, Heejune Sheen, Tianhao Wang, and Zhuoran Yang.
\newblock Training dynamics of multi-head softmax attention for in-context
  learning: Emergence, convergence, and optimality.
\newblock \emph{arXiv preprint arXiv:2402.19442}, 2024.

\bibitem[Collins et~al.(2024)Collins, Parulekar, Mokhtari, Sanghavi, and
  Shakkottai]{collins2024context}
Liam Collins, Advait Parulekar, Aryan Mokhtari, Sujay Sanghavi, and Sanjay
  Shakkottai.
\newblock In-context learning with transformers: Softmax attention adapts to
  function lipschitzness.
\newblock \emph{arXiv preprint arXiv:2402.11639}, 2024.

\bibitem[Dao \& Gu(2024)Dao and Gu]{dao2024transformers}
Tri Dao and Albert Gu.
\newblock Transformers are ssms: Generalized models and efficient algorithms
  through structured state space duality.
\newblock \emph{arXiv preprint arXiv:2405.21060}, 2024.

\bibitem[De et~al.(2024)De, Smith, Fernando, Botev, Cristian-Muraru, Gu,
  Haroun, Berrada, Chen, Srinivasan, et~al.]{de2024griffin}
Soham De, Samuel~L Smith, Anushan Fernando, Aleksandar Botev, George
  Cristian-Muraru, Albert Gu, Ruba Haroun, Leonard Berrada, Yutian Chen,
  Srivatsan Srinivasan, et~al.
\newblock Griffin: Mixing gated linear recurrences with local attention for
  efficient language models.
\newblock \emph{arXiv preprint arXiv:2402.19427}, 2024.

\bibitem[Deng et~al.(2023)Deng, Song, Xie, and Yang]{deng2023unmasking}
Yichuan Deng, Zhao Song, Shenghao Xie, and Chiwun Yang.
\newblock Unmasking transformers: A theoretical approach to data recovery via
  attention weights.
\newblock \emph{arXiv preprint arXiv:2310.12462}, 2023.

\bibitem[Deora et~al.(2023)Deora, Ghaderi, Taheri, and
  Thrampoulidis]{deora2023optimization}
Puneesh Deora, Rouzbeh Ghaderi, Hossein Taheri, and Christos Thrampoulidis.
\newblock On the optimization and generalization of multi-head attention.
\newblock \emph{arXiv preprint arXiv:2310.12680}, 2023.

\bibitem[Ding et~al.(2023)Ding, Levinboim, Wu, Goodman, and
  Soricut]{ding2023causallm}
Nan Ding, Tomer Levinboim, Jialin Wu, Sebastian Goodman, and Radu Soricut.
\newblock Causallm is not optimal for in-context learning.
\newblock \emph{arXiv preprint arXiv:2308.06912}, 2023.

\bibitem[Dun et~al.(2023)Dun, Garcia, Zheng, Awadallah, Kyrillidis, and
  Sim]{dun2023sweeping}
Chen Dun, Mirian~Hipolito Garcia, Guoqing Zheng, Ahmed~Hassan Awadallah,
  Anastasios Kyrillidis, and Robert Sim.
\newblock Sweeping heterogeneity with smart mops: Mixture of prompts for llm
  task adaptation.
\newblock \emph{arXiv preprint arXiv:2310.02842}, 2023.

\bibitem[Ergen et~al.(2022)Ergen, Neyshabur, and Mehta]{ergen2022convexifying}
Tolga Ergen, Behnam Neyshabur, and Harsh Mehta.
\newblock Convexifying transformers: Improving optimization and understanding
  of transformer networks.
\newblock \emph{arXiv:2211.11052}, 2022.

\bibitem[Fu et~al.(2022)Fu, Dao, Saab, Thomas, Rudra, and R{\'e}]{fu2022hungry}
Daniel~Y Fu, Tri Dao, Khaled~K Saab, Armin~W Thomas, Atri Rudra, and
  Christopher R{\'e}.
\newblock Hungry hungry hippos: Towards language modeling with state space
  models.
\newblock \emph{arXiv preprint arXiv:2212.14052}, 2022.

\bibitem[Fu et~al.(2023)Fu, Chen, Jia, and Sharan]{fu2023transformers}
Deqing Fu, Tianqi Chen, Robin Jia, and Vatsal Sharan.
\newblock Transformers learn higher-order optimization methods for in-context
  learning: A study with linear models.
\newblock In \emph{NeurIPS 2023 Workshop on Mathematics of Modern Machine
  Learning}, 2023.

\bibitem[Garg et~al.(2022)Garg, Tsipras, Liang, and Valiant]{garg2022can}
Shivam Garg, Dimitris Tsipras, Percy~S Liang, and Gregory Valiant.
\newblock What can transformers learn in-context? a case study of simple
  function classes.
\newblock \emph{Advances in Neural Information Processing Systems},
  35:\penalty0 30583--30598, 2022.

\bibitem[Gatmiry et~al.(2024)Gatmiry, Saunshi, Reddi, Jegelka, and
  Kumar]{gatmirycan}
Khashayar Gatmiry, Nikunj Saunshi, Sashank Reddi, Stefanie Jegelka, and Sanjiv
  Kumar.
\newblock Can looped transformers learn to implement multi-step gradient
  descent for in-context learning?
\newblock In \emph{Proceedings of the 41st International Conference on Machine
  Learning}, pp.\  15130--15152, 2024.

\bibitem[Grazzi et~al.(2024)Grazzi, Siems, Schrodi, Brox, and
  Hutter]{grazzi2024mamba}
Riccardo Grazzi, Julien Siems, Simon Schrodi, Thomas Brox, and Frank Hutter.
\newblock Is mamba capable of in-context learning?
\newblock \emph{arXiv preprint arXiv:2402.03170}, 2024.

\bibitem[Gu \& Dao(2023)Gu and Dao]{gu2023mamba}
Albert Gu and Tri Dao.
\newblock Mamba: Linear-time sequence modeling with selective state spaces.
\newblock \emph{arXiv preprint arXiv:2312.00752}, 2023.

\bibitem[Gu et~al.(2021)Gu, Goel, and R{\'e}]{gu2021efficiently}
Albert Gu, Karan Goel, and Christopher R{\'e}.
\newblock Efficiently modeling long sequences with structured state spaces.
\newblock \emph{arXiv preprint arXiv:2111.00396}, 2021.

\bibitem[Huang et~al.(2024)Huang, Liang, and Yang]{huang2024non}
Ruiquan Huang, Yingbin Liang, and Jing Yang.
\newblock Non-asymptotic convergence of training transformers for next-token
  prediction.
\newblock \emph{arXiv preprint arXiv:2409.17335}, 2024.

\bibitem[Huang et~al.(2023)Huang, Cheng, and Liang]{huang2023context}
Yu~Huang, Yuan Cheng, and Yingbin Liang.
\newblock In-context convergence of transformers.
\newblock \emph{arXiv preprint arXiv:2310.05249}, 2023.

\bibitem[Ildiz et~al.(2024)Ildiz, Huang, Li, Rawat, and Oymak]{ildiz2024self}
M~Emrullah Ildiz, Yixiao Huang, Yingcong Li, Ankit~Singh Rawat, and Samet
  Oymak.
\newblock From self-attention to markov models: Unveiling the dynamics of
  generative transformers.
\newblock \emph{arXiv preprint arXiv:2402.13512}, 2024.

\bibitem[Jelassi et~al.(2022)Jelassi, Sander, and Li]{jelassi2022vision}
Samy Jelassi, Michael~Eli Sander, and Yuanzhi Li.
\newblock Vision transformers provably learn spatial structure.
\newblock In Alice~H. Oh, Alekh Agarwal, Danielle Belgrave, and Kyunghyun Cho
  (eds.), \emph{Advances in Neural Information Processing Systems}, 2022.

\bibitem[Jelassi et~al.(2024)Jelassi, Brandfonbrener, Kakade, and
  Malach]{jelassi2024repeat}
Samy Jelassi, David Brandfonbrener, Sham~M Kakade, and Eran Malach.
\newblock Repeat after me: Transformers are better than state space models at
  copying.
\newblock \emph{arXiv preprint arXiv:2402.01032}, 2024.

\bibitem[Jeon et~al.(2024)Jeon, Lee, Lei, and Van~Roy]{jeon2024information}
Hong~Jun Jeon, Jason~D Lee, Qi~Lei, and Benjamin Van~Roy.
\newblock An information-theoretic analysis of in-context learning.
\newblock \emph{arXiv preprint arXiv:2401.15530}, 2024.

\bibitem[Julistiono et~al.(2024)Julistiono, Tarzanagh, and
  Azizan]{julistiono2024optimizing}
Aaron Alvarado~Kristanto Julistiono, Davoud~Ataee Tarzanagh, and Navid Azizan.
\newblock Optimizing attention with mirror descent: Generalized max-margin
  token selection.
\newblock \emph{arXiv preprint arXiv:2410.14581}, 2024.

\bibitem[Katharopoulos et~al.(2020)Katharopoulos, Vyas, Pappas, and
  Fleuret]{katharopoulos2020transformers}
Angelos Katharopoulos, Apoorv Vyas, Nikolaos Pappas, and Fran{\c{c}}ois
  Fleuret.
\newblock Transformers are rnns: Fast autoregressive transformers with linear
  attention.
\newblock In \emph{International conference on machine learning}, pp.\
  5156--5165. PMLR, 2020.

\bibitem[Katsch(2023)]{katsch2023gateloop}
Tobias Katsch.
\newblock Gateloop: Fully data-controlled linear recurrence for sequence
  modeling.
\newblock \emph{arXiv preprint arXiv:2311.01927}, 2023.

\bibitem[Li et~al.(2023{\natexlab{a}})Li, Wang, Liu, and
  Chen]{li2023theoretical}
Hongkang Li, Meng Wang, Sijia Liu, and Pin-Yu Chen.
\newblock A theoretical understanding of shallow vision transformers: Learning,
  generalization, and sample complexity.
\newblock \emph{arXiv preprint arXiv:2302.06015}, 2023{\natexlab{a}}.

\bibitem[Li et~al.(2023{\natexlab{b}})Li, Ildiz, Papailiopoulos, and
  Oymak]{li2023transformers}
Yingcong Li, Muhammed~Emrullah Ildiz, Dimitris Papailiopoulos, and Samet Oymak.
\newblock Transformers as algorithms: Generalization and stability in
  in-context learning.
\newblock In \emph{International Conference on Machine Learning}, pp.\
  19565--19594. PMLR, 2023{\natexlab{b}}.

\bibitem[Li et~al.(2024{\natexlab{a}})Li, Huang, Ildiz, Rawat, and
  Oymak]{li2024mechanics}
Yingcong Li, Yixiao Huang, Muhammed~E Ildiz, Ankit~Singh Rawat, and Samet
  Oymak.
\newblock Mechanics of next token prediction with self-attention.
\newblock In \emph{International Conference on Artificial Intelligence and
  Statistics}, pp.\  685--693. PMLR, 2024{\natexlab{a}}.

\bibitem[Li et~al.(2024{\natexlab{b}})Li, Rawat, and Oymak]{li2024fine}
Yingcong Li, Ankit~Singh Rawat, and Samet Oymak.
\newblock Fine-grained analysis of in-context linear estimation: Data,
  architecture, and beyond.
\newblock \emph{arXiv preprint arXiv:2407.10005}, 2024{\natexlab{b}}.

\bibitem[Lin \& Lee(2024)Lin and Lee]{lin2024dual}
Ziqian Lin and Kangwook Lee.
\newblock Dual operating modes of in-context learning.
\newblock \emph{arXiv preprint arXiv:2402.18819}, 2024.

\bibitem[Mahankali et~al.(2023)Mahankali, Hashimoto, and Ma]{mahankali2023one}
Arvind Mahankali, Tatsunori~B Hashimoto, and Tengyu Ma.
\newblock One step of gradient descent is provably the optimal in-context
  learner with one layer of linear self-attention.
\newblock \emph{arXiv preprint arXiv:2307.03576}, 2023.

\bibitem[Makkuva et~al.(2024)Makkuva, Bondaschi, Girish, Nagle, Jaggi, Kim, and
  Gastpar]{makkuva2024attention}
Ashok~Vardhan Makkuva, Marco Bondaschi, Adway Girish, Alliot Nagle, Martin
  Jaggi, Hyeji Kim, and Michael Gastpar.
\newblock Attention with markov: A framework for principled analysis of
  transformers via markov chains.
\newblock \emph{arXiv preprint arXiv:2402.04161}, 2024.

\bibitem[Min et~al.(2022)Min, Lyu, Holtzman, Artetxe, Lewis, Hajishirzi, and
  Zettlemoyer]{min2022rethinking}
Sewon Min, Xinxi Lyu, Ari Holtzman, Mikel Artetxe, Mike Lewis, Hannaneh
  Hajishirzi, and Luke Zettlemoyer.
\newblock Rethinking the role of demonstrations: What makes in-context learning
  work?
\newblock \emph{arXiv preprint arXiv:2202.12837}, 2022.

\bibitem[Orvieto et~al.(2023)Orvieto, Smith, Gu, Fernando, Gulcehre, Pascanu,
  and De]{orvieto2023resurrecting}
Antonio Orvieto, Samuel~L Smith, Albert Gu, Anushan Fernando, Caglar Gulcehre,
  Razvan Pascanu, and Soham De.
\newblock Resurrecting recurrent neural networks for long sequences.
\newblock In \emph{International Conference on Machine Learning}, pp.\
  26670--26698. PMLR, 2023.

\bibitem[Oymak et~al.(2023)Oymak, Rawat, Soltanolkotabi, and
  Thrampoulidis]{oymak2023role}
Samet Oymak, Ankit~Singh Rawat, Mahdi Soltanolkotabi, and Christos
  Thrampoulidis.
\newblock On the role of attention in prompt-tuning.
\newblock In \emph{International Conference on Machine Learning}, 2023.

\bibitem[Park et~al.(2024)Park, Park, Xiong, Lee, Cho, Oymak, Lee, and
  Papailiopoulos]{park2024can}
Jongho Park, Jaeseung Park, Zheyang Xiong, Nayoung Lee, Jaewoong Cho, Samet
  Oymak, Kangwook Lee, and Dimitris Papailiopoulos.
\newblock Can mamba learn how to learn? a comparative study on in-context
  learning tasks.
\newblock \emph{arXiv preprint arXiv:2402.04248}, 2024.

\bibitem[Peng et~al.(2023)Peng, Alcaide, Anthony, Albalak, Arcadinho, Biderman,
  Cao, Cheng, Chung, Grella, et~al.]{peng2023rwkv}
Bo~Peng, Eric Alcaide, Quentin Anthony, Alon Albalak, Samuel Arcadinho, Stella
  Biderman, Huanqi Cao, Xin Cheng, Michael Chung, Matteo Grella, et~al.
\newblock Rwkv: Reinventing rnns for the transformer era.
\newblock \emph{arXiv preprint arXiv:2305.13048}, 2023.

\bibitem[Peng et~al.(2024)Peng, Goldstein, Anthony, Albalak, Alcaide, Biderman,
  Cheah, Ferdinan, Hou, Kazienko, et~al.]{peng2024eagle}
Bo~Peng, Daniel Goldstein, Quentin Anthony, Alon Albalak, Eric Alcaide, Stella
  Biderman, Eugene Cheah, Teddy Ferdinan, Haowen Hou, Przemys{\l}aw Kazienko,
  et~al.
\newblock Eagle and finch: {RWKV} with matrix-valued states and dynamic
  recurrence.
\newblock \emph{arXiv preprint arXiv:2404.05892}, 2024.

\bibitem[Peng et~al.(2021)Peng, Pappas, Yogatama, Schwartz, Smith, and
  Kong]{peng2021random}
Hao Peng, Nikolaos Pappas, Dani Yogatama, Roy Schwartz, Noah~A Smith, and
  Lingpeng Kong.
\newblock Random feature attention.
\newblock \emph{arXiv preprint arXiv:2103.02143}, 2021.

\bibitem[Qin et~al.(2024)Qin, Yang, Sun, Shen, Li, Sun, and
  Zhong]{qin2024hgrn2}
Zhen Qin, Songlin Yang, Weixuan Sun, Xuyang Shen, Dong Li, Weigao Sun, and
  Yiran Zhong.
\newblock Hgrn2: Gated linear rnns with state expansion.
\newblock \emph{arXiv preprint arXiv:2404.07904}, 2024.

\bibitem[Sahiner et~al.(2022)Sahiner, Ergen, Ozturkler, Pauly, Mardani, and
  Pilanci]{sahiner2022unraveling}
Arda Sahiner, Tolga Ergen, Batu Ozturkler, John Pauly, Morteza Mardani, and
  Mert Pilanci.
\newblock Unraveling attention via convex duality: Analysis and interpretations
  of vision transformers.
\newblock In \emph{International Conference on Machine Learning}, pp.\
  19050--19088. PMLR, 2022.

\bibitem[Sheen et~al.(2024)Sheen, Chen, Wang, and Zhou]{sheen2024implicit}
Heejune Sheen, Siyu Chen, Tianhao Wang, and Harrison~H Zhou.
\newblock Implicit regularization of gradient flow on one-layer softmax
  attention.
\newblock \emph{arXiv preprint arXiv:2403.08699}, 2024.

\bibitem[Sieber et~al.(2024)Sieber, Alonso, Didier, Zeilinger, and
  Orvieto]{sieber2024understanding}
Jerome Sieber, Carmen~Amo Alonso, Alexandre Didier, Melanie~N Zeilinger, and
  Antonio Orvieto.
\newblock Understanding the differences in foundation models: Attention, state
  space models, and recurrent neural networks.
\newblock \emph{arXiv preprint arXiv:2405.15731}, 2024.

\bibitem[Sun et~al.(2025)Sun, Jadbabaie, and Azizan]{sun2025context}
Haoyuan Sun, Ali Jadbabaie, and Navid Azizan.
\newblock In-context learning of polynomial kernel regression in transformers
  with glu layers.
\newblock \emph{arXiv preprint arXiv:2501.18187}, 2025.

\bibitem[Sun et~al.(2023)Sun, Dong, Huang, Ma, Xia, Xue, Wang, and
  Wei]{sun2023retentive}
Yutao Sun, Li~Dong, Shaohan Huang, Shuming Ma, Yuqing Xia, Jilong Xue, Jianyong
  Wang, and Furu Wei.
\newblock Retentive network: A successor to transformer for large language
  models.
\newblock \emph{arXiv preprint arXiv:2307.08621}, 2023.

\bibitem[Sun et~al.(2024)Sun, Dong, Zhu, Huang, Wang, Ma, Zhang, Wang, and
  Wei]{sun2024you}
Yutao Sun, Li~Dong, Yi~Zhu, Shaohan Huang, Wenhui Wang, Shuming Ma, Quanlu
  Zhang, Jianyong Wang, and Furu Wei.
\newblock You only cache once: Decoder-decoder architectures for language
  models.
\newblock \emph{arXiv preprint arXiv:2405.05254}, 2024.

\bibitem[Tarzanagh et~al.(2023)Tarzanagh, Li, Thrampoulidis, and
  Oymak]{tarzanagh2023transformers}
Davoud~Ataee Tarzanagh, Yingcong Li, Christos Thrampoulidis, and Samet Oymak.
\newblock Transformers as support vector machines.
\newblock \emph{arXiv preprint arXiv:2308.16898}, 2023.

\bibitem[Tarzanagh et~al.(2024)Tarzanagh, Li, Zhang, and Oymak]{ataee2024max}
Davoud~Ataee Tarzanagh, Yingcong Li, Xuechen Zhang, and Samet Oymak.
\newblock Max-margin token selection in attention mechanism.
\newblock \emph{Advances in Neural Information Processing Systems}, 36, 2024.

\bibitem[Thrampoulidis(2024)]{thrampoulidis2024implicit}
Christos Thrampoulidis.
\newblock Implicit bias of next-token prediction.
\newblock \emph{arXiv preprint arXiv:2402.18551}, 2024.

\bibitem[Tian et~al.(2023)Tian, Wang, Zhang, Chen, and Du]{tian2023joma}
Yuandong Tian, Yiping Wang, Zhenyu Zhang, Beidi Chen, and Simon Du.
\newblock Joma: Demystifying multilayer transformers via joint dynamics of mlp
  and attention.
\newblock \emph{arXiv preprint arXiv:2310.00535}, 2023.

\bibitem[Vasudeva et~al.(2024{\natexlab{a}})Vasudeva, Deora, and
  Thrampoulidis]{vasudeva2024implicit}
Bhavya Vasudeva, Puneesh Deora, and Christos Thrampoulidis.
\newblock Implicit bias and fast convergence rates for self-attention.
\newblock \emph{arXiv preprint arXiv:2402.05738}, 2024{\natexlab{a}}.

\bibitem[Vasudeva et~al.(2024{\natexlab{b}})Vasudeva, Fu, Zhou, Kau, Huang, and
  Sharan]{vasudeva2024simplicity}
Bhavya Vasudeva, Deqing Fu, Tianyi Zhou, Elliott Kau, Youqi Huang, and Vatsal
  Sharan.
\newblock Simplicity bias of transformers to learn low sensitivity functions.
\newblock \emph{arXiv preprint arXiv:2403.06925}, 2024{\natexlab{b}}.

\bibitem[Vaswani(2017)]{vaswani2017attention}
A~Vaswani.
\newblock Attention is all you need.
\newblock \emph{Advances in Neural Information Processing Systems}, 2017.

\bibitem[Von~Oswald et~al.(2023)Von~Oswald, Niklasson, Randazzo, Sacramento,
  Mordvintsev, Zhmoginov, and Vladymyrov]{von2023transformers}
Johannes Von~Oswald, Eyvind Niklasson, Ettore Randazzo, Jo{\~a}o Sacramento,
  Alexander Mordvintsev, Andrey Zhmoginov, and Max Vladymyrov.
\newblock Transformers learn in-context by gradient descent.
\newblock In \emph{International Conference on Machine Learning}, pp.\
  35151--35174. PMLR, 2023.

\bibitem[von Oswald et~al.(2023)von Oswald, Niklasson, Schlegel, Kobayashi,
  Zucchet, Scherrer, Miller, Sandler, Vladymyrov, Pascanu,
  et~al.]{von2023uncovering}
Johannes von Oswald, Eyvind Niklasson, Maximilian Schlegel, Seijin Kobayashi,
  Nicolas Zucchet, Nino Scherrer, Nolan Miller, Mark Sandler, Max Vladymyrov,
  Razvan Pascanu, et~al.
\newblock Uncovering mesa-optimization algorithms in transformers.
\newblock \emph{arXiv preprint arXiv:2309.05858}, 2023.

\bibitem[Wang et~al.(2024{\natexlab{a}})Wang, An, Cheng, Zhou, Hwang, and
  Hsieh]{wang2024one}
Ruochen Wang, Sohyun An, Minhao Cheng, Tianyi Zhou, Sung~Ju Hwang, and Cho-Jui
  Hsieh.
\newblock One prompt is not enough: Automated construction of a
  mixture-of-expert prompts.
\newblock \emph{arXiv preprint arXiv:2407.00256}, 2024{\natexlab{a}}.

\bibitem[Wang et~al.(2024{\natexlab{b}})Wang, Wei, Hsu, and
  Lee]{wang2024transformers}
Zixuan Wang, Stanley Wei, Daniel Hsu, and Jason~D Lee.
\newblock Transformers provably learn sparse token selection while
  fully-connected nets cannot.
\newblock \emph{arXiv preprint arXiv:2406.06893}, 2024{\natexlab{b}}.

\bibitem[Wu et~al.(2023)Wu, Zou, Chen, Braverman, Gu, and Bartlett]{wu2023many}
Jingfeng Wu, Difan Zou, Zixiang Chen, Vladimir Braverman, Quanquan Gu, and
  Peter~L Bartlett.
\newblock How many pretraining tasks are needed for in-context learning of
  linear regression?
\newblock \emph{arXiv preprint arXiv:2310.08391}, 2023.

\bibitem[Xie et~al.(2022)Xie, Raghunathan, Liang, and Ma]{xieexplanation}
Sang~Michael Xie, Aditi Raghunathan, Percy Liang, and Tengyu Ma.
\newblock An explanation of in-context learning as implicit bayesian inference.
\newblock In \emph{International Conference on Learning Representations}, 2022.

\bibitem[Yang et~al.(2023)Yang, Wang, Shen, Panda, and Kim]{yang2023gated}
Songlin Yang, Bailin Wang, Yikang Shen, Rameswar Panda, and Yoon Kim.
\newblock Gated linear attention transformers with hardware-efficient training.
\newblock \emph{arXiv preprint arXiv:2312.06635}, 2023.

\bibitem[Zhang et~al.(2023)Zhang, Frei, and Bartlett]{zhang2023trained}
Ruiqi Zhang, Spencer Frei, and Peter~L Bartlett.
\newblock Trained transformers learn linear models in-context.
\newblock \emph{arXiv preprint arXiv:2306.09927}, 2023.

\bibitem[Zhang et~al.(2024)Zhang, Frei, and Bartlett]{zhang2024trained}
Ruiqi Zhang, Spencer Frei, and Peter~L Bartlett.
\newblock Trained transformers learn linear models in-context.
\newblock \emph{Journal of Machine Learning Research}, 25\penalty0
  (49):\penalty0 1--55, 2024.

\bibitem[Zhao \& Thrampoulidis(2025)Zhao and Thrampoulidis]{zhao2025geometry}
Yize Zhao and Christos Thrampoulidis.
\newblock Geometry of concepts in next-token prediction: Neural-collapse meets
  semantics.
\newblock In \emph{The Second Conference on Parsimony and Learning (Recent
  Spotlight Track)}, 2025.
\newblock URL \url{https://openreview.net/forum?id=Lt6l1woQ84}.

\bibitem[Zhao et~al.(2024)Zhao, Behnia, Vakilian, and
  Thrampoulidis]{zhao2024implicit}
Yize Zhao, Tina Behnia, Vala Vakilian, and Christos Thrampoulidis.
\newblock Implicit geometry of next-token prediction: From language sparsity
  patterns to model representations.
\newblock \emph{arXiv preprint arXiv:2408.15417}, 2024.

\bibitem[Zheng et~al.(2023)Zheng, Qiu, and Ma]{zheng2023learn}
Junhao Zheng, Shengjie Qiu, and Qianli Ma.
\newblock Learn or recall? revisiting incremental learning with pre-trained
  language models.
\newblock \emph{arXiv preprint arXiv:2312.07887}, 2023.

\bibitem[Zimerman et~al.(2024)Zimerman, Ali, and Wolf]{zimerman2024unified}
Itamar Zimerman, Ameen Ali, and Lior Wolf.
\newblock A unified implicit attention formulation for gated-linear recurrent
  sequence models.
\newblock \emph{arXiv preprint arXiv:2405.16504}, 2024.

\bibitem[Zong et~al.(2024)Zong, Shao, Lu, and Zhuang]{zong2024stock}
Chang Zong, Jian Shao, Weiming Lu, and Yueting Zhuang.
\newblock Stock movement prediction with multimodal stable fusion via gated
  cross-attention mechanism.
\newblock \emph{arXiv preprint arXiv:2406.06594}, 2024.

\end{thebibliography}
\bibliographystyle{iclr2025_conference}

\newpage
\addtocontents{toc}{\protect\setcounter{tocdepth}{3}}
\tableofcontents
\appendix
\section{Experimental setup}\label{sec:exp}
\subsection{Implementation detail}

\paragraph*{Data generation. } Consider ICL problem with input in the form of multi-task prompt as described in Section~\ref{sec:multi p}. In the experiments, we set $K=2$, dimensions $d=10$ and $p=5$, uniform context length $n_1=n_2= \bar n$ where we have $n=K\bar n$, and vary $\bar n$ from $0$ to $50$. Let $(r_1,r_2):=\left(\E[\bt_1^\top\bt]/d,\E[\bt_2^\top\bt]/d\right)$ denote the correlations between in-context tasks $\bt_1,\bt_2$ and query task $\bt$. We generate task vectors as follows:
\[
\bt_1,\bt_2\sim\Nc(0,\Iden_d), \quad\textnormal{and}\quad\bt\sim\Nc(r_1\bt_1+r_2\bt_2,(1-r_1^2-r_2^2)\Iden_d).
\]
Input features are randomly sampled $((\x_{ik})_{i=1}^{\bar n})_{k=1}^K,\x_{n+1}\sim\Nc(0,\Iden_d)$, and we have $y_{ik}=\bt_k^\top\x_{ik}$ ($\sigma=0$), $k\in\{1,2\}$ and $y_{n+1}=\bt^\top\x_{n+1}$. Additionally, delimiters $\cbb_0,\cdots,\cbb_K$ are randomly sampled from $\Nc(0,\Iden_p)$. 


\paragraph*{Implementation setting.} We train 1-layer linear attention and GLA models for solving multi-prompt ICL problem as described in Section~\ref{sec:multi p}. For GLA model, we consider sigmoid-type gating function given by scalar gating: $G(\z)=\phi(\w_g^\top\z)\onebb_{(d+p+1)\times(d+p+1)}$, or vector gating: $G(\z)=\phi(\W_g\z)\onebb^\top_{d+p+1}$ where  $\phi(z)=(1+e^{-z})^{-1}$ is the activation function. 
Note that although the theoretical results are based on the model constructions (c.f.~\eqref{att mtx res} and \eqref{att mtx res delimiter}), we do not restrict the attention weights in our implementation. We train each model for $10000$ iterations with batch size $256$ and Adam optimizer with learning rate $10^{-3}$. Similar to the previous work \citep{li2024fine}, since our study focuses on the optimization landscape, ICL problems using linear attention/GLA models are non-convex, and experiments are implemented via gradient descent, we repeat $10$ model trainings from different model initialization and data sampling (e.g., different choice of delimiters) and results are presented as the minimal test risk among those $10$ trails. Results presented have been normalized by $d$. 


\paragraph*{Experimental results.} 
%
Based on the experimental setting, we can obtain the correlation matrix and vector following Definition~\ref{corr task def}
\[
\Rb=\begin{bmatrix}
    \onebb_{\bar n}\onebb_{\bar n}^\top&\zerobb\\
    \zerobb&\onebb_{\bar n}\onebb_{\bar n}^\top
\end{bmatrix}\quad\text{and}\quad\rb=\begin{bmatrix}
    r_1\onebb_{\bar n}^\top&r_2\onebb_{\bar n}^\top
\end{bmatrix}^\top.
\]
Then dotted curves display our theoretical results derive using $\bSi=\Iden$ and $\Rb,\rb$ above. Specifically, in Figure~\ref{fig:k=2}, black dashed curves represent $\Lc^\st_\wpgd$ following \eqref{optim wpgd risk} and blues dashed curves represent $\Lc^\st_\gla$ following \eqref{optim att risk}. 
We consider scenarios where $(r_1,r_2)\in\left\{(0,1),(0.2,0.8),(0.5,0.5),(0.8,0.2)\right\}$ and results are presented in Figures \eqref{fig:0 1}, \eqref{fig:0.2 0.8}, \eqref{fig:0.5 0.5} and \eqref{fig:0.8 0.2}, respectively. 

\paragraph*{$\bullet$} \texttt{GLA-wo} achieves the worst performance among all the methods. We claim that it is due to the randomness of input tokens as discussed in Section~\ref{sec:multi p}. Thanks to the introduction of delimiters as described in \eqref{mtl prompt}, data and gating is decoupled and a task-dependent weighting is learnt. Hence, \texttt{GLA} is able to achieve comparable performance to the optimal one ($\Lc^\st_\wpgd$, red dashed). Note that \texttt{GLA-wo} performs even worse than \texttt{LinAtt}. It comes from the fact the weighting induced by \texttt{GLA-wo} varies over different input prompts and it can not implement all ones weight.  

\paragraph*{$\bullet$} The alignments between \texttt{LinAtt} (blue solid) and blue dashed curves validate our Corollary~\ref{corol:att}. In Figures~\ref{fig:0 1}, \ref{fig:0.2 0.8} and \ref{fig:0.5 0.5}, the alignments between \texttt{GLA} (red solid) and $\Lc_\wpgd$ (black dashed) verify our Theorem~\ref{thm:gla optim}, specifically, Equation~\ref{gla=wpgd}. While in \ref{fig:0.5 0.5} and \ref{fig:0.8 0.2}, \texttt{GLA} achieves the same performance as \texttt{LinAtt}. It is due to the fact that \texttt{GLA} can not weight the history higher than its present. Then the equal-weighting, e.g., $\bom=\onebb$, is the optimal weighting given such constraint. What's more, the alignment between \texttt{GLA-vector} (cyan curves) and red dashed in Figure~\ref{fig:0.8 0.2} validates our vector gating theorem in Theorem~\ref{thm:gla optim vector}.


\subsection{Multi-layer experiments}
\begin{figure}[!t]
\centering
\centering
\includegraphics[width=0.5\linewidth]{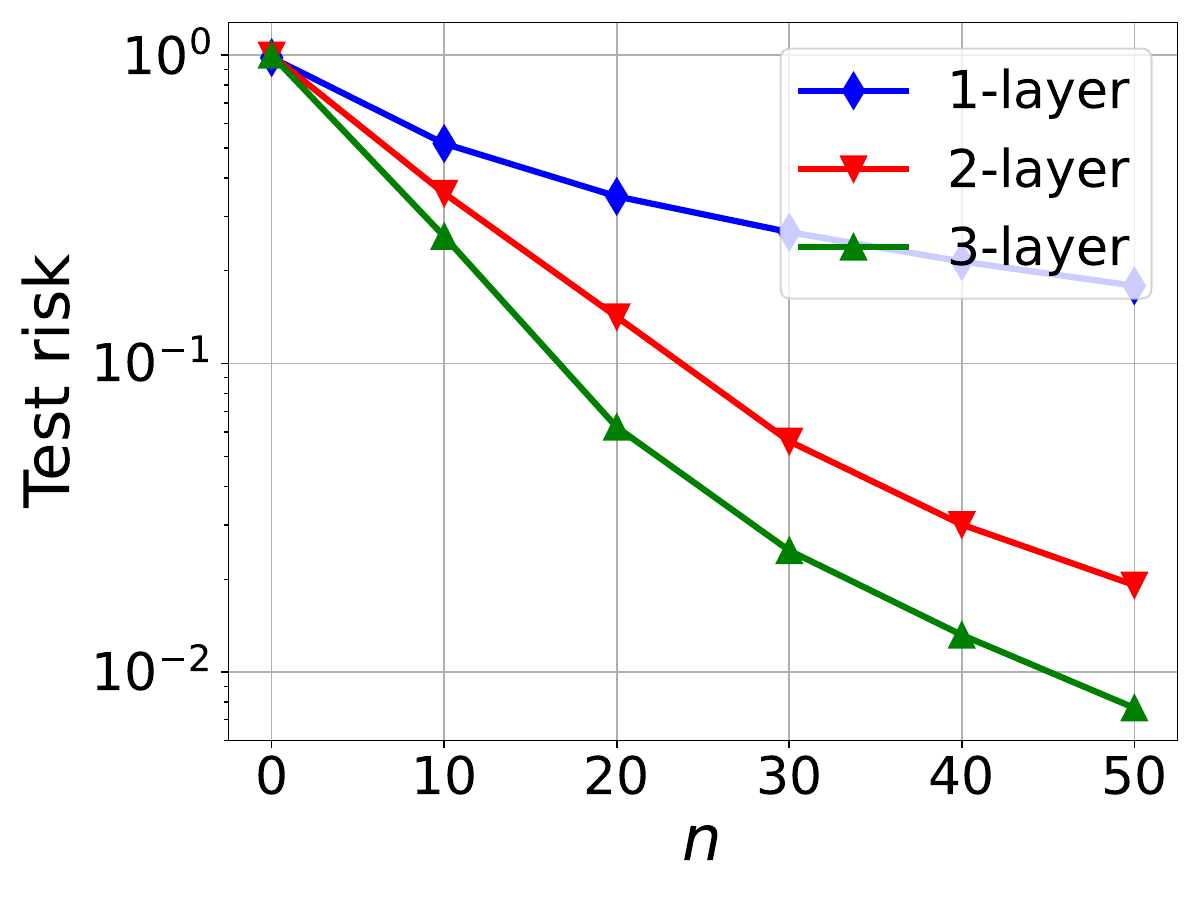}
\caption{{Multi-layer GLA experiments with $(r_1,r_2)=(0,1)$.}}\label{fig:0 1 multi}
\vspace{-10pt}
\end{figure}
In this section, we present additional experiments on multi-layer GLA models. We adopt the same experimental setup as described in Figure~\ref{fig:0 1} and Appendix~\ref{sec:exp}, with parameters setting to $(r_1, r_2) = (0, 1)$. The results are displayed in Figure~\ref{fig:0 1 multi}, where the blue, red, and green curves correspond to the performance of one-, two-, and three-layer GLA models, respectively, with the \(y\)-axis presented in log-scale. According to Theorem~\ref{lemma:multi layer}, an \(L\)-layer GLA performs \(L\) steps of WPGD, suggesting that deeper models should yield improved predictive performance. The experimental findings in Figure~\ref{fig:0 1 multi} align with the theoretical predictions of Theorem~\ref{lemma:multi layer}.

\section{GLA $\Leftrightarrow$ WPGD} \label{app:cons}
\subsection{Proof of Theroem~\ref{thm:gla=wpgd}}\label{app:1 layer}
\begin{proof}
Recap the problem settings from Section~\ref{sec:setup} where in-context samples are given by
\begin{align*}
\Z=[\z_1~\cdots~\z_n~\z_{n+1}]^\top=\begin{bmatrix}\x_1&\cdots&\x_n&\x_{n+1}\\
y_1&\cdots&y_{n}&0\end{bmatrix}^\top
\end{align*}
and let the value, key and query embeddings at time $i$ be
\[
\vb_i=\Wv^\top\z_i,\quad\kb_i=\Wk^\top\z_i,\quad\text{and}\quad\qb_i=\Wq^\top\z_i.
\]
Then we can rewrite the GLA output (cf.~\eqref{eq:GLA}) as follows:
\begin{align*}
    \ob_i=\Sb_{i}\qb_i\quad\text{and}\quad\Sb_{i}&=\Gb_{i}\odot \Sb_{i-1}+\vb_{i}\kb_{i}^\top\\
    &=\Gb_{i}\odot\cdots\Gb_1\odot\vb_1\kb_1^\top+\cdots+\Gb_i\odot\vb_{i-1}\kb_{i-1}^\top+\vb_i\kb_i^\top\\
    &=\sum_{j=1}^{i}\Gb_{j:i}\odot\vb_j\kb_j^\top,
\end{align*}
where we define
\[
\Gb_{j:i}=\Gb_{j+1}\odot\Gb_{j+2}\cdots\Gb_{i},\quad j<i,\quad\text{and}\quad\Gb_{i:i}=\onebb_{(d+1)\times (d+1)}.
\]
Consider the prediction based on the last token, then we obtain
\begin{align*}
\ob_{n+1}&=\Sb_{n+1}\qb_{n+1}\quad\text{and}\quad\Sb_{n+1}=\sum_{j=1}^{n+1}\Gb_{j:n+1}\odot\vb_j\kb_j^\top.
\end{align*}

\paragraph*{Construction 1:} Recall the model construction from \eqref{att mtx res} where
\begin{align}\Wk=\begin{bmatrix}
    \Pk&\zerobb\\
    \zerobb&0
\end{bmatrix},\quad\Wq=\begin{bmatrix}
    \Pq&\zerobb\\\zerobb&0
\end{bmatrix}\quad\text{and}\quad\Wv=\begin{bmatrix}\zerobb_{d\times d}&\zerobb\\\zerobb&1\end{bmatrix}.\label{mtx cons 1}
\end{align}
Then, given each token $\z_i=[\x_i^\top~y_i]^\top$, $i\in[n]$, single-layer GLA returns
\[
\vb_i=\begin{bmatrix}
    \zerobb\\y_i
\end{bmatrix},\quad\kb_i=\begin{bmatrix}
    \Pk^\top\x_i\\0
\end{bmatrix},\quad\text{and}\quad\qb_i=\begin{bmatrix}
    \Pq^\top\x_i\\0
\end{bmatrix},
\]
and we obtain
\[
\vb_i\kb_i^\top=\begin{bmatrix}
    \zerobb_{d\times d}&\zerobb\\
    y_i\x_i^\top\Pk&0
\end{bmatrix},\quad i\leq n,\quad\text{and}\quad\vb_{n+1}\kb_{n+1}^\top=\zerobb_{(d+1)\times(d+1)}.
\]
Therefore, since only $d$ entries in $\vb_i\kb_i^\top$ matrix are nonzero, given $\odot$ as the Hadamard product, only the corresponding $d$ entries in all $\Gb_i$ matrices are useful. Based on this observation, let 
\[
\Gb_i=\begin{bmatrix}
    *&*\\
    \g_i^\top&*
\end{bmatrix}\quad\text{and}\quad\Gb_{j:i}=\begin{bmatrix}
    *&*\\
    \g_{j:i}^\top&*,
\end{bmatrix}
\]
where $\g_{j:i}=\g_{j+1}\odot\g_{j+2}\cdots\g_i\in\R^d$ for $j<i$ and $\g_{i:i}=\onebb_d$.

Combing all together, and letting $\X,\y$ follow the definitions in \eqref{PGD loss}, we obtain
\begin{align*}
\ob_{n+1}&=\Sb_{n+1}\qb_{n+1}\\
&=\left(\sum_{j=1}^{n+1}\Gb_{j:n+1}\odot\vb_j\kb_j^\top\right)\qb_{n+1}\\
&=\begin{bmatrix}
    \zerobb_{d\times d}&\zerobb\\
    \sum_{j=1}^ny_j\x_j^\top\Pk\odot\g_{j:n+1}^\top&0
\end{bmatrix}\begin{bmatrix}
    \Pq^\top\x\\0
\end{bmatrix}\\
&=\begin{bmatrix}
    \zerobb\\
    \x^\top\Pq\left(\X\Pk\odot\bOm\right)^\top\y
\end{bmatrix}
\end{align*}
where
\[
\bOm=\begin{bmatrix}
    \g_{1:n+1}&\g_{2:n+1}&\cdots&\g_{n:n+1}
\end{bmatrix}\in\R^{n\times d}.
\]
Then if taking the last entry of $\ob_{n+1}$ as final prediction, we get
\[
f_{\gla}(\Z)=\x^\top\Pq\left(\X\Pk\odot\bOm\right)^\top\y
\]
which completes the proof of Theorem~\ref{thm:gla=wpgd}.
\end{proof}
\paragraph*{Construction 2:} Based on the construction given in \eqref{mtx cons 1}, only $d$ elements of $\Gb_i$ matrices are useful. One might ask about the effect of other entries of $\Gb_i$. Therefore, in the following, we introduce an other model construction showing that different row of $\Gb_i$ implements WPGD with different weighting. Similarly, let $\Wk,\Wq$ be the same as \eqref{mtx cons 1} but with $\Wv$ constructed by
\begin{align*}\Wv=\begin{bmatrix}\zerobb_{(d+1)\times d}&\ub\end{bmatrix}^\top\quad\text{where}\quad
\ub=[u_1~u_2~\cdots~u_{d+1}]^\top\in\R^{d+1}.
\end{align*}
Then, the value embeddings have the form of $\vb_i=y_i\ub$, which gives 
\[
\vb_i\kb_i^\top=\ub\begin{bmatrix}
    y_i\x_i^\top\Pk&\zerobb
\end{bmatrix}.
\]
Next, let 
\[
\Gb_i=\begin{bmatrix}
    (\g_{i}^1)^\top&*\\
    (\g_{i}^2)^\top&*\\
    \vdots&\vdots\\
    (\g_{i}^{d+1})^\top&*
\end{bmatrix}, \quad\text{and}\quad\Gb_{j:i}=\begin{bmatrix}
    (\g_{j:i}^1)^\top&*\\
    (\g_{j:i}^2)^\top&*\\
    \vdots&\vdots\\
    (\g_{j:i}^{d+1})^\top&*
\end{bmatrix}
,
\]
where $\g_i^{i'}\in\R^d$ corresponds to the $i'$-th row of $\Gb_i$ and $\g_{j:i}^{i'}=\g_{j+1}^{i'}\odot\g_{j+1}^{i'}\cdots\g_i^{i'}$. Then we get the output
\[
\ob_{n+1}=\begin{bmatrix}
    \sum_{j=1}^nu_1y_j\x_j^\top\Pk\odot(\g^1_{j:n+1})^\top&0\\    
    \sum_{j=1}^nu_2y_j\x_j^\top\Pk\odot(\g^2_{j:n+1})^\top&0\\
    \vdots&\\
    \sum_{j=1}^nu_{d+1}y_j\x_j^\top\Pk\odot(\g^{d+1}_{j:n+1})^\top&0

\end{bmatrix}\begin{bmatrix}
    \Pq^\top\x\\0
\end{bmatrix}=\begin{bmatrix}
    \x^\top\Pq\left(\X\Pk\odot\bOm_1\right)^\top\y\\
    \x^\top\Pq\left(\X\Pk\odot\bOm_2\right)^\top\y\\
    \vdots\\
    \x^\top\Pq\left(\X\Pk\odot\bOm_{d+1}\right)^\top\y
\end{bmatrix},
\]
where
\begin{equation}\label{eqn:allrow}
    \bOm_{i}=u_{i}\begin{bmatrix}    \g_{1:n+1}^{i}&\g_{2:n+1}^{i}&\cdots&\g_{n:n+1}^{i}
\end{bmatrix}\in\R^{n\times d},\quad i\leq d+1.
\end{equation}
Therefore, consider $(d+1)$-dimensional output $\ob_{n+1}$. Each entry implements a 1-step WPGD with same  preconditioners $\Pk,\Pq$ and different weighting matrices $\bOm$'s. The weighting matrix of $i$'th entry is determined by the $i$'th row of all gating matrices. Note that if consider the last entry of $\ob_{n+1}$ as prediction, it returns the same result as \textbf{Construction 1} above, where only last rows of $\Gb_i$'s are useful.

Additionally, suppose that the final prediction is given after a linear head $\hb$, that is, $f_{\gla}(\Z)=\hb^\top\ob_{n+1}$, and let $\hb=[h_1~h_2~\cdots~h_{d+1}]^\top\in\R^{d+1}$. Then
\begin{align}
f_{\gla}(\Z)=\hb^\top\ob_{n+1}=\x^\top\Pq\left(\X\Pk\odot\bar\bOm\right)^\top\y\label{pred d dim}
\end{align}
where
\begin{align}
\bar\bOm=\sum_{i=1}^{d+1}h_i\bOm_i=\sum_{i=1}^{d+1}h_iu_{i}\begin{bmatrix}
    \g_{1:n+1}^{i}&\g_{2:n+1}^{i}&\cdots&\g_{n:n+1}^{i}
\end{bmatrix}\in\R^{n\times d}.\label{gate d dim}
\end{align}
Then, single-layer GLA still returns one-step WPGD with updated weighting matrix. 

\subsection{Proof of Theorem~\ref{lemma:multi layer}}\label{app:multi layer}
In this section, we first prove the following theorem, which differs from Theorem~\ref{lemma:multi layer} only in that $\W_{q,\ell}$ is parameterized by $\Pql$ instead of $-\Pql$. Theorem~\ref{lemma:multi layer} can then be directly obtained by setting $\Pql$ to $-\Pql$.
\begin{theorem}
Consider an $L$-layer GLA with residual connections, where each layer follows the weight construction specified in \eqref{att mtx res}. Let $\W_{k,\ell}$ and $\W_{q,\ell}$ in the $\ell$'th layer parameterized by $\Pkl, \Pql \in \R^{d \times d}$, for $\ell \in [L]$. Let the gating be a function of the features, e.g., $\Gb_i =G(\x_i)$, and define $\bOm$ as in Theorem~\ref{thm:gla=wpgd}. 
Additionally, denote the masking as $\M_i = \begin{bmatrix}
    \Iden_i & \zerobb\\
    \zerobb & \zerobb
\end{bmatrix} \in \R^{n \times n}$, and  let $\hat\bt_0=\btm{i}{0} = \zerobb$ for all $i \in [n]$. 

{Then, the last entry of the $i$'th token output at the $\ell$'th layer ($o_{i,\ell}$) is:}
\begin{itemize}
    \item For all $i \leq n$, $ o_{i,\ell} = y_i - \x_i^\top \bt_{i,\ell}$, where $\bt_{i,\ell} = \bt_{i,\ell-1} {+\Pql} \left(\nabla_{i,\ell} \oslash \g_{i:n+1}\right)$,
    \item $ o_{n+1,\ell} = -\x^\top \hat\bt_{\ell}$, where $\hat\bt_{\ell} = (1 + \alpha_{\ell}) \hat\bt_{\ell-1} +\Pql \left(\nabla_{n,\ell} \oslash \g_{n+1}\right)$, and $\alpha_\ell = \x^\top \Pql \Pkl^\top \x$.
\end{itemize}
Here, letting $\B_{\ell} = [\btm{1}{\ell}~\cdots~\btm{n}{\ell}]^\top$, $\X_{\ell} = \X\Pb_{k,\ell} \odot \bOm$, 
and $\y_\ell = \diag{\X\B_\ell^\top}$, we define 
$$\nabla_{i,\ell} = \X_{\ell}^\top \M_i \left(\y_\ell-\y\right).$$
\end{theorem}

\begin{proof}
Suppose that the input prompt  for $\ell$'th layer is
\begin{align}
\Z_\ell=
\begin{bmatrix}
\tilde\x_1 & \dots & \tilde\x_n & \tilde\x_{n+1} \\
\tilde y_1 & \dots & \tilde y_n & \tilde y_{n+1}
\end{bmatrix}^\top
\in\R^{(n+1)\times(d+1)}\label{input ell}
\end{align}
and let $\kb_{i,\ell},\qb_{i,\ell},\vb_{i,\ell}$ be their corresponding key, query and value embeddings.
Recap the model construction from \eqref{att mtx res} and follow the same analysis in Appendix~\ref{app:1 layer}. Let $\Sb^\ell_i$, $i\in[n+1]$ be the corresponding states. We have
\begin{align*}
\Sb_i^\ell&=\sum_{j=1}^i\Gb_{j:i}^\ell\odot\vb_{j,\ell}\kb_{j,\ell}^\top\\
&=\begin{bmatrix}
    \zerobb_{d\times d}&\zerobb\\
    \sum_{j=1}^i\tilde y_j\tilde\x_j^\top\Pkl\odot(\g_{j:i}^\ell)^\top&0
\end{bmatrix}
\end{align*}
where
\[
\Gb_i^\ell=G(\tilde\x_i)=\begin{bmatrix}
    *&*\\
    (\g_i^\ell)^\top&*
\end{bmatrix}\quad\text{and}\quad\Gb_{j:i}^\ell=\begin{bmatrix}
    *&*\\
    (\g_{j:i}^\ell)^\top&*
\end{bmatrix}.
\]
%
%
Additionally, recap that we have $\M_i=\begin{bmatrix}
    \Iden_i&\zerobb\\\zerobb&\zerobb
\end{bmatrix}$ and $\oslash$ denotes Hadamard division.  Let
\[
\bOm^\ell=\begin{bmatrix}
    \g_{i:n+1}^\ell&\cdots&\g_{n:n+1}^\ell
\end{bmatrix}.
\]
Then, defining 
\[
\tilde\X=[\tilde\x_1~~\tilde\x_2~~\cdots~~\tilde\x_{n}]^\top\in\R^{n\times d}\quad\text{and}\quad\tilde\y=[\tilde y_1~~\tilde y_2~~\cdots~~\tilde y_n]^\top\in\R^n,
\]
and letting $\ob_{i,\ell}$, $i\in[n+1]$ be their outputs, we get: 
\begin{itemize}
    \item For $i\leq n$,
\begin{align*}
\sum_{j=1}^i\tilde y_j\Pkl^\top\tilde\x_j\odot\g_{j:i}^\ell&=\left(\sum_{j=1}^i\tilde y_j\Pkl^\top\tilde\x_j\odot\g_{j:n+1}^\ell\right)\oslash\g_{i:n+1}^\ell\\
&=(\tilde\X\Pkl\odot\bOm^\ell)^\top\M_i\tilde\y\oslash\g_{i:n+1}^\ell,
\end{align*}
and
\begin{align}
\ob_{i,\ell}&=\Sb_{i}^\ell\qb_{i,\ell}=\begin{bmatrix}
    \zerobb_{d\times d}&\zerobb\\
    \left((\Xt\Pkl\odot\bOm^\ell)^\top\M_i\tilde\y\oslash\g^\ell_{i:n+1}\right)^\top&0
\end{bmatrix}\begin{bmatrix}
    \Pql^\top\tilde\x_i\\0
\end{bmatrix}\nn\\
&=\begin{bmatrix}
    \zerobb\\
    \tilde\x_i^\top\Pql\left((\Xt\Pkl\odot\bOm^\ell)^\top\M_i\tilde\y\oslash\g_{i:n+1}^\ell\right)
\end{bmatrix}.\label{output}
\end{align}
\item For $i=n+1$,
\begin{align*}
\sum_{j=1}^{n+1}\tilde y_j\Pkl^\top\tilde\x_j\odot\g_{j:i}^\ell&=(\tilde\X\Pkl\odot\bOm^\ell)^\top\tilde\y+\tilde y_{n+1}\Pkl^\top\tilde\x_{n+1},
\end{align*}
and
\begin{align}
\ob_{n+1,\ell}&=\begin{bmatrix}
    \zerobb_{d\times d}&\zerobb\\
    \left((\tilde\X\Pkl\odot\bOm^\ell)^\top\tilde\y+\tilde y_{n+1}\Pkl^\top\tilde\x_{n+1}\right)^\top&0
\end{bmatrix}\begin{bmatrix}
    \Pql^\top\tilde\x_{n+1}\\0
\end{bmatrix}\nn\\
&=\begin{bmatrix}
    \zerobb\\
    \tilde\x_{n+1}^\top\Pql\left((\tilde\X\Pkl\odot\bOm^\ell)^\top\tilde\y+\tilde y_{n+1}\Pkl^\top\tilde\x_{n+1}\right)
\end{bmatrix}\nn\\
&=\begin{bmatrix}
    \zerobb\\
    \tilde y_{n+1}\tilde\x_{n+1}^\top\Pql\Pkl^\top\tilde\x_{n+1}+\tilde\x_{n+1}^\top\Pql\left((\tilde\X\Pkl\odot\bOm^\ell)^\top\tilde\y\right).
\end{bmatrix}\label{output last}
\end{align}
\end{itemize}
From \eqref{output} and \eqref{output last}, the first $d$ entries of all the ($d+1$)-dimensional outputs (e.g, $\ob_{i,\ell}$, $i\in[n+1]$) are zero. Given the multi-layer GLA as formulated in \eqref{gla multi} where residual connections are applied, we have that the first $d$ dimension of all the input tokens remain the same. That is, $\tilde\x_i\equiv\x_i$. Additionally, since gating only depends on the feature $\x_i$, then all the layers return the same gatings (e.g, $\Gb_i^\ell\equiv\Gb_i$).

Note that if the gating function varies across layers or depends on the entire token rather than just the feature $\x_i$, each layer will have its own gating, leading to $\Gb_i^\ell\neq\Gb_i$. In this theorem, we assume identical gating matrices across layers for simplicity and clarity. However, an extended version of this theorem, without this assumption, can be directly derived.

Therefore, we obtain
\begin{equation*}
\begin{split}
&\ob_{i,\ell}=\begin{bmatrix}
    \zerobb\\
    \x_i^\top\Pql\left(\X_\ell^\top\M_i\tilde\y\oslash\g_{i:n+1}\right)
\end{bmatrix},\quad i\leq n\quad\\
&\ob_{n+1,\ell}=\begin{bmatrix}
    \zerobb\\
    \tilde y_{n+1}\x^\top\Pql\Pkl^\top\x+\x^\top\Pql\left(\X_\ell^\top\tilde\y\oslash\g_{n+1}\right)
\end{bmatrix}.
\end{split}
\end{equation*}
where we define $\X_\ell:=\X\Pkl\odot\bOm$. 
%


Since we focus on the last/$(d+1)$'th entry of each token output, we have
\begin{equation}\label{output y}
\begin{split}
&o_{i,\ell}=
    \x_i^\top\Pql\left(\X_\ell^\top\M_i\tilde\y\oslash\g_{i:n+1}\right),\quad i\leq n,\\
    &o_{n+1,\ell}=
    \tilde y_{n+1}\x^\top\Pql\Pkl^\top\x+\x^\top\Pql\left(\X_\ell^\top\tilde\y\oslash\g_{n+1}\right).
\end{split}
\end{equation}
It remains to get the $\tilde y_i$ for each layer's input. Let inputs of 
$(\ell-1)$'th and $\ell$'th layer be
\begin{align*}
\Z_{\ell-1}=
\begin{bmatrix}
\x_1 & \dots & \x_n & \x \\
y_1^{\ell-1} & \dots & y_n^{\ell-1} & y^{\ell-1}
\end{bmatrix}\quad\text{and}\quad\Z_\ell=\begin{bmatrix}
\x_1 & \dots & \x_n & \x \\
y_1^{\ell} & \dots & y_n^{\ell} & y^{\ell}
\end{bmatrix}.
\end{align*}
Define 
\[
\y^{\ell-1}:=[y_1^{\ell-1}~~\cdots~~y_n^{\ell-1}]^\top\in\R^n.
\]
\begin{itemize}
    \item For $i\leq n$, from \eqref{output y} and \eqref{gla multi}, we have
\begin{align*}
&y_i^\ell=y_i^{\ell-1}+\x_i^\top\Pql\left(\X_\ell^\top\M_i\y^{\ell-1}\oslash\g_{i:n+1}\right)   \\
\Longrightarrow~&y_i^{\ell}-y_i^{\ell-1}=\x_i^\top\Pql\left(\X_\ell^\top\M_i\y^{\ell-1}\oslash\g_{i:n+1}\right). 
\end{align*}
Suppose that $y_i-y_i^\ell=\x_i^\top\bt_i^\ell$. Then
\begin{align}
    \y^{\ell}=\y-[\x_1^\top\bt_1^{\ell}~~\cdots~~\x_n^\top\bt_n^{\ell}]^\top=\y-\diag{\X\B^\top}\label{y ell}
\end{align} 
and
\begin{align*}
&y_i-\x_i^\top\bt_i^{\ell}-(y_i-\x_i^\top\bt_i^{\ell-1})=\x_i^\top\Pql\left(\X_\ell^\top\M_i(\y-\diag{\X\B^\top})\oslash\g_{i:n+1}\right)\\
\Longrightarrow&\x_i^\top\bt_i^\ell=\x_i^\top\bt_i^{\ell-1}-\x_i^\top\Pql\left(\X_\ell^\top\M_i(\y-\diag{\X\B^\top})\oslash\g_{i:n+1}\right).
\end{align*}
Hence,
\begin{align}
    y_i^\ell=y_i-\x_i^\top\bt_i^\ell\quad\text{where}\quad\bt_i^\ell=\bt_i^{\ell-1}-\Pql\left(\X_\ell^\top\M_i(\y-\diag{\X\B^\top})\oslash\g_{i:n+1}\right)\label{final y i}
\end{align}
\item For $i=n+1$, from \eqref{output y} and \eqref{gla multi}, we have 
\[
y^{\ell}=y^{\ell-1}+ y^{\ell-1}\x^\top\Pql\Pkl^\top\x+\x^\top\Pql\left(\X_\ell^\top\y^{\ell-1}\oslash\g_{n+1}\right).
\]
Setting $\alpha_\ell=\x^\top\Pql\Pkl^\top\x$ and recalling $\y^\ell$ from \eqref{y ell}, we get
\[
y^{\ell}=(1+\alpha_\ell)y^{\ell-1}+\x^\top\Pql\left(\X_\ell^\top(\y-\diag{\X\B^\top}))\oslash\g_{n+1}\right).
\]
Additionally, let $y^\ell=-\x^\top\bt^\ell$. Then,
\[
-\x^\top\bt^{\ell}=-(1+\alpha_\ell)\x^\top\bt^{\ell-1}+\x^\top\Pql\left(\X_\ell^\top(\y-\diag{\X\B^\top}))\oslash\g_{n+1}\right).
\]
Hence,
\begin{align}
y^\ell=-\x^\top\bt^\ell\quad\text{where}\quad\bt^\ell=(1+\alpha_\ell)\bt^{\ell-1}-\Pql\left(\X_\ell^\top(\y-\diag{\X\B^\top}))\oslash\g_{n+1}\right)\label{final y n+1}
\end{align}
\end{itemize}
Combining \eqref{final y i} and \eqref{final y n+1} together completes the proof.
\end{proof}

\section{Optimization landscape of WPGD}
We first provide the following Lemma.

\begin{lemma}[Existence of Fixed Point] \label{lem:exist:fixedpoint}
Consider the functions $h_1(\bar{\gamma})$ and $h_2(\gamma)$ defined in \eqref{func:h} and \eqref{func:g}, respectively. The composite function  $ h_1(h_2(\gamma))+1$ has at least one fixed point \(\gamma^\star \geq 1\), i.e., there exists \(\gamma^\star \geq 1\) such that \(
 h_1(h_2(\gamma^\star)) +1 = \gamma^\star\).
\end{lemma}

\begin{proof}
We prove existence using continuity and boundedness arguments along with the intermediate value theorem. Note that \(h_1(\bar{\gamma})\) and \(h_2(\gamma)\) are continuous functions. Thus, their composition \(f(\gamma) := h_1(h_2(\gamma))+1\) is continuous for all \(\gamma \geq 1\).

We examine the behavior of \(f(\gamma)\) at the boundaries of its domain: At \(\gamma = 1\), we have $  h_2(1) = c_0 > 0$, where \(c_0\) is a positive constant. Therefore, $ f(1) = h_1(h_2(1))+1 = h_1(c_0)+1 > 0$,  which is finite and positive since \(h_1\) is positive for all non-negative inputs.   As \(\gamma \to \infty\),    $ h_2(\gamma) \to c_\infty > 0$,  where \(c_\infty\) is a positive constant.

For sufficiently large \(\gamma\), we can establish that \(f(\gamma) < \gamma\). To see this, note that we have shown that \(h_2(\gamma) \to c_\infty\) as \(\gamma \to \infty\), where \(c_\infty\) is a fixed positive constant. Further, for any fixed positive value \(\bar{\gamma}\), the function \(h_1(\bar{\gamma})\) is bounded, i.e., $h_1(\bar{\gamma}) \leq 
    \max_{1 \leq i \leq n} \lambda_i$, which follows from the fact that \(h_1(\bar{\gamma})\) is a weighted average of the \(\lambda_i\) values.  Therefore, there exists a constant \(K \) such that \(f(\gamma) \leq K\) for all \(\gamma \geq 1\). Hence, it follows that for all \(\gamma > K\), we have \(f(\gamma) \leq K < \gamma\).

Now, consider the function \(\bar{f}(\gamma) = f(\gamma) - \gamma\), which is continuous on \([1,\infty)\) since both \(f(\gamma)\) and \(\gamma\) are continuous. We have  \(\bar{f}(1) = f(1) - 1 = h_1(c_0) > 0\) and for some \(\gamma_1 > K\), we have \(\bar{f}(\gamma_1) = f(\gamma_1) - \gamma_1 < 0\).  Since \(\bar{f}\) is continuous and changes sign from positive to negative on the interval \([1, \gamma_1]\), by the Intermediate Value Theorem, there exists at least one \(\gamma^\star \in (1, \gamma_1)\) such that \(\bar{f}(\gamma^\star) = 0\). This implies \(f(\gamma^\star) - \gamma^\star = 0\), or equivalently, \(f(\gamma^\star) = \gamma^\star\).
Therefore, there exists at least one fixed point \(\gamma^\star \geq 1\) for the function \(h_1(h_2(\gamma))+1\).
\end{proof}

\subsection{Proof of Theorem~\ref{thm:station:wpgd}}

\begin{proof}
Recapping the objective from \eqref{eqn:wpgd:risk} and following Definitions~\ref{corr task def} and \ref{def:multitask}, we have
\begin{align}
\Lc(\Pb,\bom)&=\E\left[\left(y-\x^\top \Pb \X(\bom \odot \y)\right)^2\right]\nn\\
&=\E\left[y^2\right]-2\E\left[y\x^\top\Pb\X(\bom\odot\y)\right]+\E\left[\left(\x^\top\Pb\X(\bom\odot\y)\right)^2\right].\nn
\end{align}
Let $y = \x^\top\bt + \xi$ and $y_i = \x_i^\top\bt_i + \xi_i$, for all  $i \in [n]$, where $\xi, \xi_i \sim \Nc(0, \sigma^2)$ are i.i.d. Then,
\[
\E[y^2]=\E[(\x^\top\bt+\xi)^2]=\tr{\bSi}+\sigma^2,
\]
and
\begin{align*}
\E\left[y\x^\top\Pb\X(\bom\odot\y)\right]&=\E\left[(\bt^\top\x+\xi)\x^\top\Pb\sum_{i=1}^n\omega_i\x_i(\x_i^\top\bt_i+\xi_i)\right]\\
    &=\E\left[\bt^\top\x\x^\top\Pb\sum_{i=1}^n\omega_i\x_i\x_i^\top\bt_i\right]\\
    &=\tr{\bSi\Pb\bSi\sum_{i=1}^n\omega_i\E\left[\bt_i\bt^\top\right]}\\
    &=\tr{\bSi^2\Pb}\bom^\top\rb.
\end{align*}
Here, the last equality comes from the fact that since $\bt_i-r_{ij}\bt_j$ is independent of $\bt_j$ for $i,j\in[n+1]$ following Definition~\ref{corr task def}, we have $\E\left[\bt_i\bt^\top\right]=r_{i,n+1}\Iden_d$ and $\sum_{i=1}^n\omega_i\E\left[\bt_i\bt^\top\right]$ returns $\bom^\top\rb\cdot\Iden_d$. 

Hence, 
\begin{align*}
\E\left[\left(\x^\top\Pb\X(\bom\odot\y)\right)^2\right]&=\E\left[\x^\top\Pb\left(\sum_{i=1}^n\omega_i(\x_i^\top\bt_i+\xi_i)\x_i\right)\left(\sum_{i=1}^n\omega_i\x_i^\top(\x_i^\top\bt_i+\xi_i)\right)\Pb^\top\x\right]\\
&=
\tr{\Pb^\top\bSi\Pb\E\left[\sum_{i=1}^n\omega_i^2(\x_i^\top\bt_i+\xi_i)^2\x_i\x_i^\top\right]} \\
&+ 
 \tr{\Pb^\top\bSi\Pb\E\left[\sum_{i\neq j}\omega_i\omega_j(\x_i^\top\bt_i+\xi_i)\x_i\x_j^\top(\x_j^\top\bt_j+\xi_j)\right]},
\end{align*}
where
\begin{align*}
\tr{\Pb^\top\bSi\Pb\E\left[\sum_{i=1}^n\omega_i^2(\x_i^\top\bt_i+\xi_i)^2\x_i\x_i^\top\right]}&=\tr{\Pb^\top\bSi\Pb\E\left[\sum_{i=1}^n\omega_i^2(\x_i^\top\bt_i\bt_i^\top\x_i+\sigma^2)\x_i\x_i^\top\right]}\\
&=\tn{\bom}^2\tr{\Pb^\top\bSi\Pb\left(\E\left[\x\x^\top\x\x^\top\right]+\sigma^2\bSi\right)}\\
&=\tn{\bom}^2\left(\tr{\bSi\Pb^\top\bSi\Pb}\left(\tr{\bSi}+\sigma^2\right)+2\tr{\bSi^2\Pb^\top\bSi\Pb}\right),
\end{align*}
and
\begin{align*}
    \tr{\Pb^\top\bSi\Pb\E\left[\sum_{i\neq j}\omega_i\omega_j(\x_i^\top\bt_i+\xi_i)\x_i\x_j^\top(\x_j^\top\bt_j+\xi_j)\right]}&=\tr{\Pb^\top\bSi\Pb\E\left[\sum_{i\neq j}\omega_i\omega_j\x_i\x_i^\top\bt_i\bt_j^\top\x_j\x_j^\top\right]}\\
    &=\tr{\Pb^\top\bSi\Pb}\sum_{i\neq j}\omega_i\omega_j\E[\bt_i^\top\bt_j]\\
    &=\tr{\bSi^2\Pb^\top\bSi\Pb}\bom^\top(\Rb-\Iden)\bom.
\end{align*}
Combining all together and letting $M:=\tr{\bSi}+\sigma^2$, we obtain
\begin{align}
\nonumber 
    \Lc(\Pb,\bom)&=M-2\tr{\bSi^2\Pb}\bom^\top\rb 
    \\
    &+M\tn{\bom}^2\tr{\bSi\Pb^\top\bSi\Pb}+(\tn{\bom}^2+\bom^\top\Rb\bom)\tr{\bSi^2\Pb^\top\bSi\Pb}.\label{app:loss}
\end{align}
For simplicity, and without loss of generality, let 
\begin{align}\label{eqn:pt}
 \Pt = \sqrt{\bSi} \Pb \sqrt{\bSi}.    
\end{align}
Then, we obtain
\begin{equation}
    \begin{split}
    \Lc(\Pt,\bom)&=M-2\tr{\bSi\Pt}\bom^\top\rb \\
     &+M\tn{\bom}^2\tr{\Pt^\top\Pt}+(\tn{\bom}^2+\bom^\top\Rb\bom)\tr{\bSi\Pt^\top\Pt}. 
    \end{split}
\end{equation}
Further, the gradients can be written as
\begin{align}
&\nabla_{\Pt}\Lc(\Pt,\bom)=-2\bom^\top\rb\bSi+2M\tn{\bom}^2\Pt+2(\tn{\bom}^2+\bom^\top\Rb\bom)\bSi\Pt,\label{grad P}\\
&\nabla_{\bom}\Lc(\Pt,\bom)=-2\tr{\bSi\Pt}\rb+2M\tr{\Pt^\top\Pt}\bom+2\tr{\bSi\Pt^\top\Pt}(\Iden_n+\Rb)\bom.\label{grad w}
\end{align}
Using the first-order optimality condition, and setting  $\nabla_{\Pt}\Lc(\Pt,\bom)=0$ and $\nabla_{\bom}\Lc(\Pt,\bom)=0$, we obtain
\begin{subequations}
\begin{equation}\label{eqn:p:opt1} 
    \begin{split}
      \Pt&=\left(M\tn{\bom}^2\Iden+(\tn{\bom}^2+\bom^\top\Rb\bom)\bSi\right)^{-1}\bSi\bom^\top\rb\\
    &=\frac{\bom^\top\rb}{M\tn{\bom}^2}\left(\frac{\tn{\bom}^2+\bom^\top\Rb\bom}{M\tn{\bom}^2}\Iden+\bSi^{-1}\right)^{-1}\\
    &=\frac{\bom^\top\rb}{M\tn{\bom}^2}\left(\frac{\gamma}{M}\cdot\Iden+\bSi^{-1}\right)^{-1},    
    \end{split}
\end{equation}    
where 
\begin{align*}
\gamma:=\frac{\bom^\top\Rb\bom}{\tn{\bom}^2}+1.
\end{align*}
Further,
\begin{equation}\label{eqn:w:opt1}
\begin{split}
\bom&=\left(\left(M\tr{\Pt^\top\Pt}+\tr{\bSi\Pt^\top\Pt}\right)\Iden+\tr{\bSi\Pt^\top\Pt}\Rb\right)^{-1}\tr{\bSi\Pt}\rb\\
    &=\frac{\tr{\bSi\Pt}}{M\tr{\Pt^\top\Pt}+\tr{\bSi\Pt^\top\Pt}}\left(\Iden+\frac{\tr{\bSi\Pt^\top\Pt}}{M\tr{\Pt^\top\Pt}+\tr{\bSi\Pt^\top\Pt}}\Rb\right)^{-1}\rb.   
\end{split}
\end{equation}
\end{subequations}
Let 
\begin{align*}
 \bSi_{\gamma} := \frac{\gamma}{M}\cdot\Iden + \bSi^{-1}.  
\end{align*}
Then, we get
\begin{align*}
\frac{\tr{\bSi\Pt^\top\Pt}}{M\tr{\Pt^\top\Pt}+\tr{\bSi\Pt^\top\Pt}}&=\left(1+M\frac{\tr{\Pt^\top\Pt}}{\tr{\bSi\Pt^\top\Pt}}\right)^{-1}\\
    &=\left(1+M\frac{\tr{\bSi_{\gamma}^{-2}}}{\tr{\bSi\bSi_{\gamma}^{-2}}}\right)^{-1}\\
    &=\left(1+M\sum_{i=1}^d\frac{s_i^2}{(M+\gamma s_i)^2}\left(\sum_{i=1}^d\frac{s_i^3}{(M+\gamma s_i)^2}\right)^{-1}\right)^{-1} \\
    &=:h_2(\gamma).
\end{align*}
Here, the last equality follows from  eigen decomposition $\bSi=\Ub \textnormal{diag}(\mathbf{s})\Ub^\top$ with  $\s = [s_1, \ldots, s_d]^\top \in \R^d_{++}$.

Now, plugging $     \Pt$ defined in \eqref{eqn:p:opt1} within $\bom$ given in \eqref{eqn:w:opt1}, we obtain
\begin{align}\label{eqn:w:opt2}
\bom = \frac{\tr{\bSi\Pt}}{M\tr{\Pt^\top\Pt}+\tr{\bSi\Pt^\top\Pt}} \cdot\left(h_2(\gamma) \cdot \Rb+\Iden\right)^{-1}\rb.
\end{align}
Using the above formulae for $\bom$, we rewrite  $\gamma=\bom^\top\Rb\bom/\tn{\bom}^2+1$  as
\begin{equation}\label{eqn:comps1}
\begin{split}
    \gamma-1 &= \frac{\rb^\top (  h_2(\gamma) \Rb+\Iden)^{-1}\Rb(  h_2(\gamma) \Rb+\Iden)^{-1}\rb}{ \rb^\top (  h_2(\gamma) \Rb+\Iden)^{-2}\rb}\\
&= \sum_{i=1}^n \frac{\la_ia_i^2}{ \left(1+  h_2(\gamma) \la_i\right)^2} \left(\sum_{i=1}^n \frac{a_i^2}{\left(1+h_2(\gamma)  \la_i\right)^2}\right)^{-1}\\
&=: h_1(h_2(\gamma)),  
\end{split}    
\end{equation}
where the second equality follows from Assumption~\ref{assum:rb:in:eigenspace} where $\rb=\Eb\ab$ with $\ab=[a_1,\cdots,a_n]^\top\in\R^n$, and the fact that $\Rb = \Eb \text{diag}(\bla) \Eb^\top$ denotes the eigen decomposition of $\Rb$, with $\bla = [\lambda_1, \ldots, \lambda_n]^\top \in \R^n_{+}$.

It follows from Lemma~\ref{lem:exist:fixedpoint} that there exists \(\gamma^\star \geq 1\) such that \(
 h_1(h_2(\gamma^\star)) +1= \gamma^\star\). From \eqref{eqn:p:opt1} and \eqref{eqn:w:opt2}, we obtain 
\begin{equation}
    \begin{split}
&\Pt=C(\rb, \bom, \bSi) \cdot \left(\frac{\gamma^\star}{M}\cdot\Iden+\bSi^{-1}\right)^{-1}, \quad \textnormal{and} \quad  \\   &\bom =  c(\rb, \bom, \bSi) \cdot\left(h_2(\gamma^\star) \cdot \Rb+\Iden\right)^{-1}\rb.
\end{split}
\end{equation}    
for some $C(\rb, \bom, \bSi) = \frac{\bom^\top\rb}{M\tn{\bom}^2}$ and $c(\rb, \bom, \bSi) =\frac{\tr{\bSi\Pt}}{M\tr{\Pt^\top\Pt}+\tr{\bSi\Pt^\top\Pt}}$.  

Now, using the our definition  $\Pt = \sqrt{\bSi} \Pb \sqrt{\bSi} $, we obtain
\begin{align*}
\Pb(\gamma^\st) &= C(\rb, \bom, \bSi) \cdot \bSi^{-\frac{1}{2}} \left( \frac{\gamma^\star}{M} \cdot \bSi + \Iden \right)^{-1}\bSi^{-\frac{1}{2}}, \quad \textnormal{and} \quad \\ \bom(\gamma^\st) &= c(\rb, \bom, \bSi) \cdot \Big(h_2(\gamma^\star) \cdot \Rb + \Iden \Big)^{-1} \rb.  
\end{align*}    
This completes the proof. 
\end{proof}

\subsection{Proof of Theorem~\ref{thm:unique:wpgd}}\label{sec:app:wpgd}

We first provide the following Lemma.

\begin{lemma}\label{lem:fixedpoint}
Consider the functions $h_1(\bar{\gamma})$ and $h_2(\gamma)$ defined in \eqref{func:h} and \eqref{func:g}, respectively,  where \(\lambda_i \geq 0\), \(a_i \neq 0\) for at least one \(i\), \(s_i > 0\), and \(M = \sigma^2 + \sum_{i=1}^{d}s_i > 0\).
Suppose $\Delta_{\bSi}\cdot \Delta_{\Rb} <M+ s_{\min}$,  where \( \Delta_{\bSi} \) and \( \Delta_{\Rb} \) denote the effective spectral gaps of \( \bSi \) and \( \Rb \), respectively, as given in  \eqref{eqn:spec:gap};  and \( s_{\min} \) is the smallest eigenvalue of \( \bSi \).
We have that
\begin{align*}
 \left|\frac{\partial h_2}{\partial \gamma} \cdot \frac{\partial h_1}{\partial h_2}\right| & 
\leq \frac{\Delta_{\bSi}^2 \cdot \Delta_{\Rb}^2}{(M+s_{\min})^2} <1.
\end{align*} 
\end{lemma}

\begin{proof}
Let
\[
B(\gamma) = \sum_{i=1}^d \frac{s_i^3}{ \left(M + \gamma s_i \right)^2}, \quad C(\gamma) = \sum_{i=1}^d \frac{s_i^2}{ \left(M + \gamma s_i \right)^2}, \quad A(\gamma) = 1 + M \frac{C(\gamma)}{B(\gamma)}.
\]
The derivatives of \( B(\gamma) \) and \( C(\gamma) \) are
\[
B'(\gamma) = -2 \sum_{i=1}^d \frac{s_i^4}{ \left(M + \gamma s_i \right)^3}, \quad C'(\gamma) = -2 \sum_{i=1}^d \frac{s_i^3}{ \left(M + \gamma s_i \right)^3}.
\]
The gradient of \( h_2(\gamma)=A(\gamma)^{-1} \) is
\begin{equation}\label{eqn:partial:gg}
\frac{\partial h_2}{\partial \gamma} = - M \left(   \frac{1}{A(\gamma) B(\gamma)}\right)^2 \left( C'(\gamma) B(\gamma) - C(\gamma) B'(\gamma)\right).
\end{equation}
It can be seen that 
\begin{equation*}
    \begin{split}
 \left(\frac{1}{A(\gamma)}\right)^2&\leq M^{-2}\left(\sum_{i=1}^d \frac{s_i^3}{ \left(M + \gamma s_i \right)^2} \right)^{2} \left(\sum_{i=1}^d \frac{s_i^2}{  \left(M+ \gamma s_i \right)^2}\right)^{-2}, 
    \end{split}
\end{equation*}
which implies that 
\begin{subequations}
\begin{equation}\label{eqn:partial:g:parta}
\left(   \frac{1 }{A(\gamma) B(\gamma)}\right)^2 \leq M^{-2}  \left(\sum_{i=1}^d \frac{s_i^2}{  \left(M + \gamma s_i \right)^2}\right)^{-2}.
\end{equation}
Further, we have
\begin{equation}\label{eqn:partial:g:partb}
\begin{split}
 C'(\gamma) B(\gamma) - C(\gamma) B'(\gamma) &=-2\sum_{i=1}^d\frac{s_i^3}{(M+\gamma s_i)^3}\sum_{i=1}^d\frac{s_i^3}{(M+\gamma s_i)^2}\\
 &+2\sum_{i=1}^d\frac{s_i^2}{(M+\gamma s_i)^2}\sum_{i=1}^d\frac{s_i^4}{(M+\gamma s_i)^3}\\
    &=\sum_{i=1}^d\sum_{j=1}^d\frac{ T_{ij} }{(M+\gamma s_i)^3(M+\gamma s_j)^3}\\
    &= M\cdot \sum_{i=1}^d\sum_{j=1}^d\frac{s_i^2s_j^2(s_i-s_i)^2}{(M+\gamma s_i)^3(M+\gamma s_j)^3}, 
\end{split}
\end{equation}
where 
\begin{equation}
\begin{split}
  T_{ij} &= s_i^2(M+\gamma s_i)s_j^4+s_i^4s_j^2(M+\gamma s_j ) \\
  &-s_i^3s_j^3(M+\gamma s_j)-s_i^3(M+\gamma s_i)s_j^3 \\
  & = s_i^2s_j^2 \left(M\cdot(s_j^2+s_i^2-2s_is_j)+\gamma(s_is_j^2+s_i^2s_j-s_is_j^2-s_i^2s_j)\right).
\end{split}
\end{equation}
\end{subequations}
Thus,  substituting \eqref{eqn:partial:g:parta} and \eqref{eqn:partial:g:partb} into \eqref{eqn:partial:gg}, we obtain
\begin{equation}
\left|\frac{\partial h_2}{\partial \gamma} \right| \leq M\cdot M^{-1} \left(\sum_{i=1}^d \frac{s_i^2}{  \left(M+ \gamma s_i \right)^2}\right)^{-2} \sum_{i,j=1}^d \frac{s_i^2 s_j^2 (s_i - s_j)^2}{(M+\gamma s_i)^3(M+\gamma s_j)^3}.
\end{equation}
Next, we derive $     \frac{\partial h_1}{\partial \bar{\gamma}}$. Let
\[
D(\bar{\gamma}) = \sum_{i=1}^n \frac{\lambda_i a_i^2}{(1+\bar{\gamma}\lambda_i)^2},  \quad E(\bar{\gamma}) = \sum_{i=1}^n \frac{a_i^2}{(1+\bar{\gamma}\lambda_i)^2}.
\]
We have 
\[
D'(\bar{\gamma}) = -2 \sum_{i=1}^n \frac{\lambda_i^2 a_i^2}{(1+\bar{\gamma}\lambda_i)^3}, \quad E'(\bar{\gamma}) = -2 \sum_{i=1}^n \frac{\lambda_i a_i^2}{(1+\bar{\gamma}\lambda_i)^3}.
\]
The derivative of \( h_1 \) with respect to \( \bar{\gamma} \) is given by
\begin{equation}\label{eqn:partialh}
     \frac{\partial h_1}{\partial \bar{\gamma}} =-\left(\frac{1}{E(\bar{\gamma})}\right)^2 \left( E(\bar{\gamma})D'(\bar{\gamma}) - D(\bar{\gamma})E'(\bar{\gamma})\right).
\end{equation}
Substituting into \eqref{eqn:partialh}, we get
\begin{subequations}
\begin{equation}\label{eqn:partial:h:parta}
\left(\frac{1}{E(\bar{\gamma})}\right)^2=   \left( \sum_{i=1}^n \frac{a_i^2}{(1 + \bar{\gamma} \lambda_i)^2} \right)^{-2},
\end{equation}
and
\begin{equation}\label{eqn:partial:h:partb}
\begin{split}
E(\bar{\gamma})D'(\bar{\gamma}) - D(\bar{\gamma})E'(\bar{\gamma}) &=  2 \sum_{i=1}^n \frac{\lambda_i^2 a_i^2}{(1 + \bar{\gamma} \lambda_i)^3} \sum_{i=1}^n \frac{a_i^2}{(1 + \bar{\gamma} \lambda_i)^2}  \\
&- 2  \sum_{i=1}^n \frac{\lambda_i a_i^2}{(1 + \bar{\gamma} \lambda_i)^2} \sum_{i=1}^n \frac{a_i^2 \lambda_i}{(1 + \bar{\gamma} \lambda_i)^3} \\
&=\sum_{i=1}^n\sum_{j=1}^n\frac{\bar{T}_{ij}}{(1 + \bar{\gamma} \lambda_i)^3(1 + \bar{\gamma} \lambda_j)^3}\\
&=\sum_{i=1}^n\sum_{j=1}^n\frac{a_i^2a_j^2 \left(\lambda_i^2+\lambda_j^2-2\lambda_i\lambda_j\right)}{(1 + \bar{\gamma} \lambda_i)^3(1 + \bar{\gamma} \lambda_j)^3}.   
\end{split}
\end{equation}
Here,  
\begin{equation}
\begin{split}
\bar{T}_{ij}&=\lambda_i^2a_i^2a_j^2(1+\bar{\gamma}\lambda_j)+a_i^2(1+\bar{\gamma}\lambda_i)\lambda_j^2a_j^2\\
&-\lambda_ia_i^2(1+\bar{\gamma}\lambda_i)a_j^2\lambda_j-a_i^2\lambda_i\lambda_ja_j^2(1+\bar{\gamma}\lambda_j) \\
&=a_i^2a_j^2 \left(\lambda_i^2(1+\bar{\gamma}\lambda_j)+(1+\bar{\gamma}\lambda_i)\lambda_j^2-\lambda_i(1+\bar{\gamma}\lambda_i)\lambda_j-\lambda_i\lambda_j(1+\bar{\gamma}\lambda_j)\right).
\end{split}
\end{equation}
\end{subequations}
Hence, substituting \eqref{eqn:partial:h:parta} and \eqref{eqn:partial:h:partb} into \eqref{eqn:partialh} gives 
\begin{align}\label{eqn:partialh_final}
     \frac{\partial h_1}{\partial \bar{\gamma}} = -  \left( \sum_{i=1}^n \frac{a_i^2}{(1 + \bar{\gamma} \lambda_i)^2} \right)^{-2} \sum_{i,j=1}^n \frac{a_i^2a_j^2 (\lambda_i -\lambda_j)^2 }{(1 + \bar{\gamma} \lambda_i)^3(1 + \bar{\gamma} \lambda_j)^3}.
\end{align}

Now, for the combined derivative, we have
\begin{align*}
\left|\frac{\partial h_2}{\partial \gamma} \cdot \frac{\partial h_1}{\partial \bar{\gamma}}\right| &\leq   \left(\sum_{i =1 }^d \frac{s_i^2}{ \left(M+\gamma s_i \right)^2}\right)^{-2} \sum_{i,j=1}^d \frac{ s_i^2 s_j^2 (s_i - s_j)^2}{ \left(M+\gamma s_i \right)^3 \left(M+\gamma s_j\right)^3} \\
& \cdot  \left( \sum_{i=1}^n \frac{a_i^2}{(1 + \bar{\gamma} \lambda_i)^2} \right)^{-2} \sum_{i,j=1}^n \frac{a_i^2a_j^2 (\lambda_i -\lambda_j)^2}{(1 + \bar{\gamma} \lambda_i)^3 (1 + \bar{\gamma} \lambda_j)^3}.
\end{align*}
Note that $M+\gamma s_i $ and $1 + \bar{\gamma} \lambda_j$ are nonnegative for all $i,j$. Hence, 
\begin{equation*}
  \begin{split}
 \left|\frac{\partial h_2}{\partial \gamma} \cdot \frac{\partial h_1}{\partial \bar{\gamma}}\right| 
& \leq \left(\sum_{i =1 }^d \frac{s_i^2 \left(M+\gamma s_i \right)}{  \left(M+ \gamma s_i \right)^3}\right)^{-2} \\
&\cdot \sum_{i,j=1}^d \frac{s_i^2 s_j^2 \cdot \Delta_1  \cdot   \left(M+ \gamma s_j\right)\left(M+ \gamma s_i\right)  }{ \left(M+\gamma s_i \right)^3  \left(M+\gamma s_j\right)^3} \\
&\cdot  \left( \sum_{i=1}^n \frac{a_i^2(1+\bar{\gamma}\lambda_i)}{(1 + \bar{\gamma} \lambda_i)^3} \right)^{-2} \\
&\cdot \sum_{i, j =1}^d \frac{a_i^2a_j^2\cdot \Delta_2 \cdot (1 + \bar{\gamma} \lambda_i) \left(1 + \bar{\gamma} \lambda_j\right) }{(1 + \bar{\gamma} \lambda_i)^3 \left(1 + \bar{\gamma} \lambda_j\right)^3}, 
\end{split}  
\end{equation*}
where
\begin{equation}
\Delta_1 := \max_{i, j \in  \mathcal{S} }  \frac{(s_i - s_j)^2}{\left(M+ \gamma s_j\right)\left(M+ \gamma s_i\right)},    \quad \Delta_2:= \max_{i, j \in  \mathcal{S}}  \frac{(\lambda_i- \lambda_j)^2}{(1 + \bar{\gamma} \lambda_i)(1+\bar{\gamma}\lambda_j)}.
\end{equation}
Here, $\mathcal{S} =\{i \in [n]~|~\lambda_i \neq 0\} \subset [n]$.

Finally, setting $\bar{\gamma}=h_2(\gamma)$, since $\gamma\geq1$ and by our assumption $\Delta_{\bSi}\cdot \Delta_{\Rb} <M+ s_{\min}$, we obtain
\begin{align*}
| h_1'(h_2(\gamma)) \cdot  h_2'(\gamma) | = \left|\frac{\partial h_2}{\partial \gamma} \cdot \frac{\partial h_1}{\partial h_2}\right| & \leq \Delta_1 \cdot \Delta_2 
\leq \frac{\Delta_{\bSi}^2 \cdot \Delta_{\Rb}^2}{(M+s_{\min})^2} <1.
\end{align*}
where \( \Delta_{\bSi} \) and \( \Delta_{\Rb} \) are the spectral gaps of \( \bSi \) and \( \Rb \);  and \( s_{\min} \) is the smallest eigenvalue of \( \bSi \); and  $M= \sigma^2 + \sum_{i=1}^d s_i$. 
\end{proof}

\begin{proof}[Proof of Theorem~\ref{thm:unique:wpgd}]
It follows from Lemma~\ref{lem:exist:fixedpoint} that there exists \(\gamma^\star \geq 1\) such that \(
 h_1(h_2(\gamma^\star)) +1 = \gamma^\star\). Further,  Lemma \ref{lem:fixedpoint} establishes that the mapping $h_1(h_2(\gamma))+1$ is a contraction mapping, since it satisfies $|\partial h_1(h_2(\gamma))/\partial \gamma|<1$. Therefore, by the Banach fixed-point theorem, there exists a \textit{unique} fixed-point solution, denoted by $\gamma^\star$, satisfying
$
\gamma^\star = h_1(h_2(\gamma^\star))+1.
$
This completes the proof of statement \ref{item:fixedpoint}. 

Next, we establish \ref{item:globalmin}. First, from Theorem \ref{thm:station:wpgd}, the stationary point $(\Pb^\star, \bom^\star)$, up to rescaling, is defined as
\begin{align*}
    \Pb^\star &= \bSi^{-\frac{1}{2}} \left( \frac{\gamma^\star }{M} \cdot \bSi + \Iden \right)^{-1} \bSi^{-\frac{1}{2}} \quad \textnormal{and} \quad \bom^\star = \left( h_2(\gamma^\star) \cdot \Rb + \Iden \right)^{-1} \rb,
\end{align*}
where $\gamma^\star$ is a fixed point of composite function \( h_1(h_2(\gamma))+1\).

Given the coercive nature of the loss function $\Lc(\Pb,\bom)$,  a global minimum exists. Since the global minimum must satisfy first-order optimality conditions, it must lie within the set of stationary points characterized by Theorem~\ref{thm:station:wpgd}.

Now, consider arbitrary global minimizers $(\hat{\Pb},\hat{\bom})$. Define scaling factors $\alpha, \beta > 0$ and write $(\hat{\Pb}, \hat{\bom})=(\alpha \Pb^\star, \beta \bom^\star)$. Substituting this scaling into the loss function gives:
\begin{align}
\nonumber
\Lc(\alpha \Pb^\star, \beta \bom^\star) &= M - 2\alpha\beta \tr{\bSi^2\Pb^\star}{\bom^\star}^\top \rb+ M\alpha^2\beta^2\|\bom^\star\|^2\tr{\bSi {\Pb^\star}^\top \bSi \Pb^\star} \\
&+ \alpha^2\beta^2(\|\bom^\star\|^2 + {\bom^\star}^\top \Rb \bom^\star)\tr{\bSi^2 {\Pb^\star}^\top \bSi \Pb^\star}.    
\end{align}
Differentiating this expression with respect to $\alpha$ and $\beta$ and setting the derivatives equal to zero, we obtain the equation
\begin{subequations}\label{eqn:ABC}
\begin{align}
 -2A + 2\alpha\beta (MCB + (C+D)E)=0,   
\end{align}
where we define
\begin{align}
\nonumber
 A&:=\tr{\bSi^2 \Pb^\star}{\bom^\star}^\top \rb, \quad B:=\tr{\bSi{\Pb^\star}^\top \bSi \Pb^\star},\\
 C&:=\|\bom^\star\|^2, \quad D:={\bom^\star}^\top \Rb\bom^\star, \quad E:=\tr{\bSi^2{\Pb^\star}^\top \bSi \Pb^\star}.   
\end{align}
\end{subequations}
Equation~\eqref{eqn:ABC} implies a unique constraint on the product $\alpha \beta$, ensuring no independent rescaling beyond a fixed ratio can further minimize the loss. Hence, no distinct minimizers exist other than scaling the original $(\Pb^\star,\bom^\star)$ pair. Thus, the stationary point $(\Pb^\star, \bom^\star)$ from Theorem \ref{thm:station:wpgd} is indeed a unique minimizer up to rescaling.
\end{proof}

\subsection{Proof of Corollary~\ref{coroll:unique}}
\begin{proof}
Since by assumption $\bSi = \Iden$, it follows from \eqref{func:g} that
\begin{align*}
   h_2(\gamma^\star) &= \left(1 + (d + \sigma^2) \sum_{i=1}^d \frac{1}{(d + \sigma^2 + \gamma^\star + 1)^2} \left(\sum_{i=1}^d \frac{1}{(d + \sigma^2 + \gamma^\star + 1)^2}\right)^{-1}\right)^{-1} \\
   &= \frac{1}{d + \sigma^2 + 1}. 
\end{align*}
Substituting this into \eqref{eqn:pstar_and_wstar} gives
\[
\Pb^\star = \Iden, \quad \textnormal{and} \quad 
\bom^\star = \left( \Rb + (d + \sigma^2 + 1)\Iden \right)^{-1} \rb.
\]
Now, recall that the objective function is given by
\[
\Lc(\bom) = M - 2 \tr{\bSi^2 \Pb} \bom^\top \rb + M\tn{\bom}^2 \tr{\bSi \Pb^\top \bSi \Pb} + (\tn{\bom}^2 + \bom^\top \Rb \bom) \tr{\bSi^2 \Pb^\top \bSi \Pb},
\]
and, by assumption, \( M = \sigma^2 + d \). 

Substituting $\Pb^\star = \Iden$ and $\bom^\star = \left( \Rb + (d + \sigma^2 + 1)\Iden \right)^{-1} \rb$ into the objective \eqref{app:loss}, and using $\bSi = \Iden$, we get:
\[
\Lc(\bom^\star) = (\sigma^2 + d) - 2 \cdot d \cdot \rb^\top \bom^\star + (\sigma^2 + d) \cdot \|{\bom^\st}\|^2d + d \left( \|\bom^\star\|^2 + {\bom^\st}^\top \Rb \bom^\star \right).
\]
The expression simplifies as
\[
\Lc(\bom^\star) = (\sigma^2 + d) - 2d \cdot \rb^\top \left( \Rb + (d + \sigma^2 + 1)\Iden \right)^{-1} \rb + (\sigma^2 + d) d\|{\bom^\st}\|^2 + d \left( \|\bom^\star\|^2 + \bom^\star{}^\top \Rb \bom^\star \right).
\]

Next, we compute \(\|\bom^\star\|^2\) and \({\bom^\st}^\top \Rb \bom^\star\). By definition, we have
\[
\|\bom^\star\|^2 = \rb^\top \left( \Rb + (d + \sigma^2 + 1)\Iden \right)^{-2} \rb,
\]
and
\[
\bom^\star{}^\top \Rb \bom^\star = \rb^\top \left( \Rb + (d + \sigma^2 + 1)\Iden \right)^{-1} \Rb \left( \Rb + (d + \sigma^2 + 1)\Iden \right)^{-1} \rb.
\]
Thus,
\begin{align*}
(d+\sigma^2+1)\|\bom^\star\|^2 + \bom^\star{}^\top \Rb \bom^\star &= \rb^\top \left( \Rb + (d + \sigma^2 + 1)\Iden \right)^{-1} \left( (d+\sigma^2+1)\Iden + \Rb \right) \left( \Rb + (d + \sigma^2 + 1)\Iden \right)^{-1} \rb\\
&=\rb^\top\left( \Rb + (d + \sigma^2 + 1)\Iden \right)^{-1} \rb.
\end{align*}
Substituting this result back into the objective function gives
\[
\Lc(\bom^\star) = (\sigma^2 + d) -  d \cdot \rb^\top\left( \Rb + (d + \sigma^2 + 1)\Iden \right)^{-1} \rb .
\]
\end{proof}

\section{Loss landscape of 1-layer GLA}\label{app:gla optim}
\subsection{Proof of Theorem~\ref{thm:gla optim}}\label{app:proof gla optim}

\begin{proof}

We first prove that under Assumption~\ref{assum delimiter gate}, $\Lc_\gla^\st=\min_{\Pb\in\R^{d\times d},\bom\in\Wc}~\Lc_\wpgd(\Pb,\bom)$ where $\Wc$ is the search space of weighting vector $\bom\in\R^{n}$ defined as
\[
\Wc:=\left\{\left[\omega_1\onebb_{n_1}^\top~\cdots~\omega_K\onebb_{n_K}^\top\right]^\top\in\R^{n}~\Big|~0\leq\omega_i\leq\omega_j\leq1,~\forall 1\leq i\leq j\leq K\right\}.
\]

\paragraph*{Step 1:} Define a set $\bar\Wc:=\left\{\left[\omega_1~\cdots~\omega_n\right]^\top\in\R^{n}~\Big|~0\leq\omega_i\leq\omega_j\leq1,~\forall 1\leq i\leq j\leq n\right\}$ and we have $\Wc\in\bar\Wc$. Given scalar gating $\Gb_i=\begin{bmatrix}*&*\\g_i\onebb^\top&*\end{bmatrix}$, following \eqref{wpgd y}, the weighting vector returns
\[
\bom:=[g_{1:n+1}~\cdots~g_{n:n+1}]^\top.
\]
Since GLA with scalar gating valued in $[0,1]$ following Assumption~\ref{assum delimiter gate}, that is, $g_i\in[0,1]$, we have $g_{i:n+1}\leq g_{j:n+1}$ for $1\leq i\leq j\leq n$. Therefore, any weighting vector $\bom$ implemented by GLA gating should be inside $\bar\Wc$.

\paragraph*{Step 2:} Next, we will show that 
\[
\bom^\st\in\Wc\quad\text{where}\quad\bom^\st=\arg\min_{\Pb,\bom\in\bar\Wc}\Lc_{\wpgd}(\Pb,\bom).
\]
Define the weighting vector $\bom=[\bom_1^\top~\cdots~\bom_K^\top]^\top\in\R^n$ where we have $\bom_k=[\omega_1^{(k)}~\cdots~\omega_{n_k}^{(k)}]^\top\in\R^{n_k}$. For any $\bom\not\in\Wc$, there exist $(i,j,k)$ with $i=j-1$ such that $\omega_i^{(k)}<\omega_j^{(k)}$. Given gradient in \eqref{grad w}, we have that 
\begin{align*}
&\nabla_{\omega_i^{(k)}}\Lc-\nabla_{\omega_j^{(k)}}\Lc=2\left(M\tr{\Pt^\top\Pt}+\tr{\bSi\Pt^\top\Pt}\right)(\omega_i^{(k)}-\omega_j^{(k)}).
\end{align*}
Here, \eqref{eqn:p:opt1} gives that (optimal) $\Pt\neq 0$ and therefore, we have that $\left(M\tr{\Pt^\top\Pt}+\tr{\bSi\Pt^\top\Pt}\right)>0$ and 
$\nabla_{\omega_i^{(k)}}\Lc<\nabla_{\omega_j^{(k)}}\Lc$. Therefore either increasing $\omega_i^{(k)}$ (if $\nabla_{\omega_i^{(k)}}\Lc<0$) or decreasing 
$\omega_j^{(k)}$ (if $\nabla_{\omega_j^{(k)}}\Lc>0$) will reduce the loss. This results in showing that the optimal weighting vector $\bom^\star$ satisfies $\omega_i^{(k)}=\omega_j^{(k)}$ for any $i,j\in[n_k]$ and $k\in[K]$. Hence,  $\bom^\st\in\Wc$.

\paragraph*{Step 3:} Finally, we will show that any $\bom\in\Wc$ can be obtained via the GLA gating. Let $\bom=[\omega_1\onebb_{n_1}^\top~\cdots~\omega_K\onebb_{n_K}^\top]^\top$ be any vector in $\Wc$ and assume that $\omega_K=\alpha<1$ without loss of generality. Then such sample weighting can be achieved via the gating 
\[
\begin{bmatrix}\onebb_{n_1}^\top &\frac{\omega_1}{\omega_{1:K}}&\cdots&\onebb_{n_k}^\top&\frac{\omega_k}{\omega_{k:K}}&\cdots&\onebb_{n_K}^\top& \frac{\omega_K}{\omega_{K:K}}\end{bmatrix}^\top.
\]
Let $\omega'_k:=\frac{\omega_k}{\omega_{k:K}}$ and let $\w_g$ be in the form of
\[
\w_g=\begin{bmatrix}
    \zerobb_{d+1}\\\tilde\w_g
\end{bmatrix}\in\R^{d+p+1}.
\] Then it remains to show that there exists $\tilde\w_g$ satisfying:
\begin{align*}
    \phi(\tilde\w_g^\top\cbb_k)\begin{cases}
        =1,&k=0\\
        =\omega_k',& k\in[K]
    \end{cases}
\end{align*}
The linear independence assumption in Assumption~\ref{assum delimiter gate} implies the feasible of the problem, which completes the proof of \eqref{gla=wpgd W}.

\paragraph*{Proof of \eqref{gla=wpgd}:}Recap the optimal weighting from \eqref{eqn:pstar_and_wstar} which takes the form of
\[
\bom^\star = \left( h_2(\gamma^\star) \cdot \Rb + \Iden \right)^{-1} \rb,
\]
where $h_2(\gamma^\st)>0$. Since Assumption~\ref{assum order} holds and $n_1=n_2=\cdots=n_K:=\bar n$, $\Rb$ is block diagonal matrix, where each block is an all-ones matrix. That is 
\[
\Rb=\begin{bmatrix}
    \onebb_{\bar n\times\bar n}&\zerobb&\cdots&\zerobb\\
    \zerobb&\onebb_{\bar n\times\bar n}&\cdots&\zerobb\\
    \vdots&\vdots&\ddots&\vdots\\
    \zerobb&\zerobb&\cdots&\onebb_{\bar n\times\bar n}
\end{bmatrix}.
\]
Therefore, we get
\[
\left( h_2(\gamma^\star) \cdot \Rb + \Iden \right)^{-1}=\Iden-\frac{h_2(\gamma^\st)}{h_2(\gamma^\st)\cdot\bar n+1}\Rb.
\]
Since $\Rb\rb=\bar n\rb$, then 
\[
\bom^\st=\frac{1}{h_2(\gamma^\st)\cdot\bar n+1}\rb.
\]
Therefore, the optimal weighting (up to a scalar) is inside the set $\Wc$. Combining it with \eqref{gla=wpgd W} completes the proof.




\end{proof}

\subsection{Proof of Theorem~\ref{thm:gla optim vector}}\label{app:gla optim vector}
\begin{proof}
Following the similar proof of Theorem~\ref{thm:gla optim}, and letting 
$
\tilde\Wc:=\left\{\left[\omega_1\onebb_{n_1}^\top~\cdots~\omega_K\onebb_{n_K}^\top\right]^\top\in\R^{n}\right\}
$, we obtain
\[
\min_{\Pb\in\R^{d\times d},\bom\in\tilde\Wc}\Lc_{\wpgd}(\Pb,\bom)=\min_{\Pb\in\R^{d\times d},\bom\in\R^n}\Lc_{\wpgd}(\Pb,\bom).
\]
Therefore, it remains to show that any $\bom\in\tilde\Wc$ can be implemented via some gating function. Let $\bom=[\omega_1\onebb_{n_1}^\top~\cdots~\omega_K\onebb_{n_K}^\top]^\top$ be arbitrary weighting in $\tilde\Wc$. Theorem~\ref{thm:gla optim} has shown that if $\omega_1\leq\omega_2\leq\cdots\leq\omega_K$, GLA with scalar function can implement such increasing weighting. 

Now, inspired from Appendix~\ref{app:cons}, \textbf{Construction 2} and \eqref{gate d dim} that all dimensions in the output implement individual WPGD, the weighting can be a composition of up to $d+1$ different weights ($\ub$ is $(d+1)$-dimensional).
Therefore, for any $\omega_1,\cdots,\omega_K\in\R$, we can get $K$ separate weighting:
\begin{align*}
    &\bom_1=\omega_1[\onebb_{n_1}^\top~\cdots~\onebb_{n_K}^\top]^\top\\
    &\bom_2=(\omega_2-\omega_1)[\zerobb_{n_1}^\top~\onebb_{n_2}^\top~\cdots~\onebb_{n_K}^\top]^\top\\
    &\bom_3=(\omega_3-\omega_2)[\zerobb_{n_1}^\top~\zerobb_{n_2}^\top~\onebb_{n_3}^\top~\cdots~\onebb_{n_K}^\top]^\top\\
    &\cdots\\
    &\bom_K=(\omega_K-\omega_{K-1})[\zerobb_{n_1}^\top~\zerobb_{n_2}^\top~\zerobb_{n_3}^\top~\cdots~\zerobb_{n_{K-1}}^\top\onebb_{n_K}^\top]^\top.
\end{align*}
%
Recap from Appendix~\ref{app:cons} and \eqref{gate d dim}, and consider the construction $\tilde\W_v=[\zerobb_{(d+p+1)\times d}~~\ub~~\zerobb_{(d+p+1)\times p}]^\top$. Assumption~\ref{assum delimiter gate} implies that $K\leq p<d+p+1$.


From \eqref{pred d dim} and \eqref{gate d dim}, let $i$'th dimension implements the weighting $\bom_i$  for $i\in[K]$. Specifically, let $i$'th row of each gating matrix $\Gb_i$ implement weighting $[\zerobb_{n_1}^\top~\cdots~\zerobb_{n_{i-1}}^\top~\onebb_{n_i}^\top~\cdots~\onebb_{n_k}^\top]$ (which is feasible due to Theorem~\ref{thm:gla optim}) and set $u_ih_i=\omega_i-\omega_{i-1}$ with $\omega_0=0$. Then the composed weighting following \eqref{gate d dim} returns $\bom=\sum_{k=1}^K\bom_k$, which completes the proof.
\end{proof}

\subsection{Proof of Corollary~\ref{corol:att}}
\begin{proof}Recap from \eqref{eqn:p:opt1} that given $\bSi=\Iden$ and $\bom=\onebb$, 
\begin{align*}
      \Pb^\st&=\left((d+\sigma^2)n\Iden+(n+\onebb^\top\Rb\onebb)\Iden\right)^{-1}\onebb^\top\rb\\
      &=\frac{\onebb^\top\rb}{n(d+\sigma^2+1)+\onebb^\top\Rb\onebb}\Iden:=c\Iden.
\end{align*}  
Then taking it back to the loss function (c.f.~\eqref{app:loss}) obtains
\begin{align*}
\Lc(\Pb^\st,\bom=1)&=d+\sigma^2-2cd\onebb^\top\rb+(d+\sigma^2)c^2nd+(n+\onebb^\top\Rb\onebb)c^2d\\
&=d+\sigma^2-cd\onebb^\top\rb.
\end{align*}
It completes the proof.
\end{proof}

\end{document}